\documentclass[10pt,twocolumn,letterpaper]{article}

\usepackage[pagenumbers]{cvpr} % To force page numbers, e.g. for an arXiv version

\usepackage{multirow}
\usepackage{tabularx}
\usepackage{float}
\usepackage{tcolorbox}
\usepackage{listings}
\usepackage{colortbl}
\usepackage[normalem]{ulem}
\useunder{\uline}{\ul}{}
\usepackage{fontawesome5}
\usepackage{amsmath}
\usepackage{makecell}
\usepackage{microtype}

\definecolor{cvprblue}{rgb}{0.21,0.49,0.74}
\usepackage[pagebackref,breaklinks,colorlinks,allcolors=cvprblue]{hyperref}

\def\paperID{xxxx} % *** Enter the Paper ID here
\def\confName{CVPR}
\def\confYear{2026}

\definecolor{color_integrate}{HTML}{d6b8ff}
\definecolor{color_grounding}{HTML}{faa472}
\definecolor{color_recognition}{HTML}{7b9eff}
\definecolor{color_spatial}{HTML}{b0e76e}
\definecolor{color_complex_scene}{HTML}{79d8d8}

\newcommand{\tagrec}[1]{\textcolor{color_recognition}{\textbf{#1}}}
\newcommand{\tagspatial}[1]{\textcolor{color_spatial}{\textbf{#1}}}
\newcommand{\taggnd}[1]{\textcolor{color_grounding}{\textbf{#1}}}
\newcommand{\tagcomplex}[1]{\textcolor{color_complex_scene}{\textbf{#1}}}

\newcommand{\tabrec}[1]{\textcolor{color_recognition}{#1}}
\newcommand{\tabspatial}[1]{\textcolor{color_spatial}{#1}}
\newcommand{\tabgnd}[1]{\textcolor{color_grounding}{#1}}

\title{MOAT: Evaluating LMMs for \textcolor{color_integrate}{Capability Integration} and \textcolor{color_grounding}{Instruction Grounding}}

\author{Zhoutong Ye, Mingze Sun, Huan-ang Gao, Xutong Wang, Xiangyang Wang, Yu Mei, Chang Liu \\ Qinwei Li, Chengwen Zhang, Qinghuan Lan, Chun Yu, Yuanchun Shi \\ \\
 Department of Computer Science and Technology, Tsinghua University
\\
{\tt\small yezt24@mails.tsinghua.edu.cn},
{\tt\small chunyu@tsinghua.edu.cn}
}

\begin{document}
\twocolumn[{
    \maketitle
    \begin{center}
    \captionsetup{type=figure}
    \includegraphics[width=0.98\textwidth]{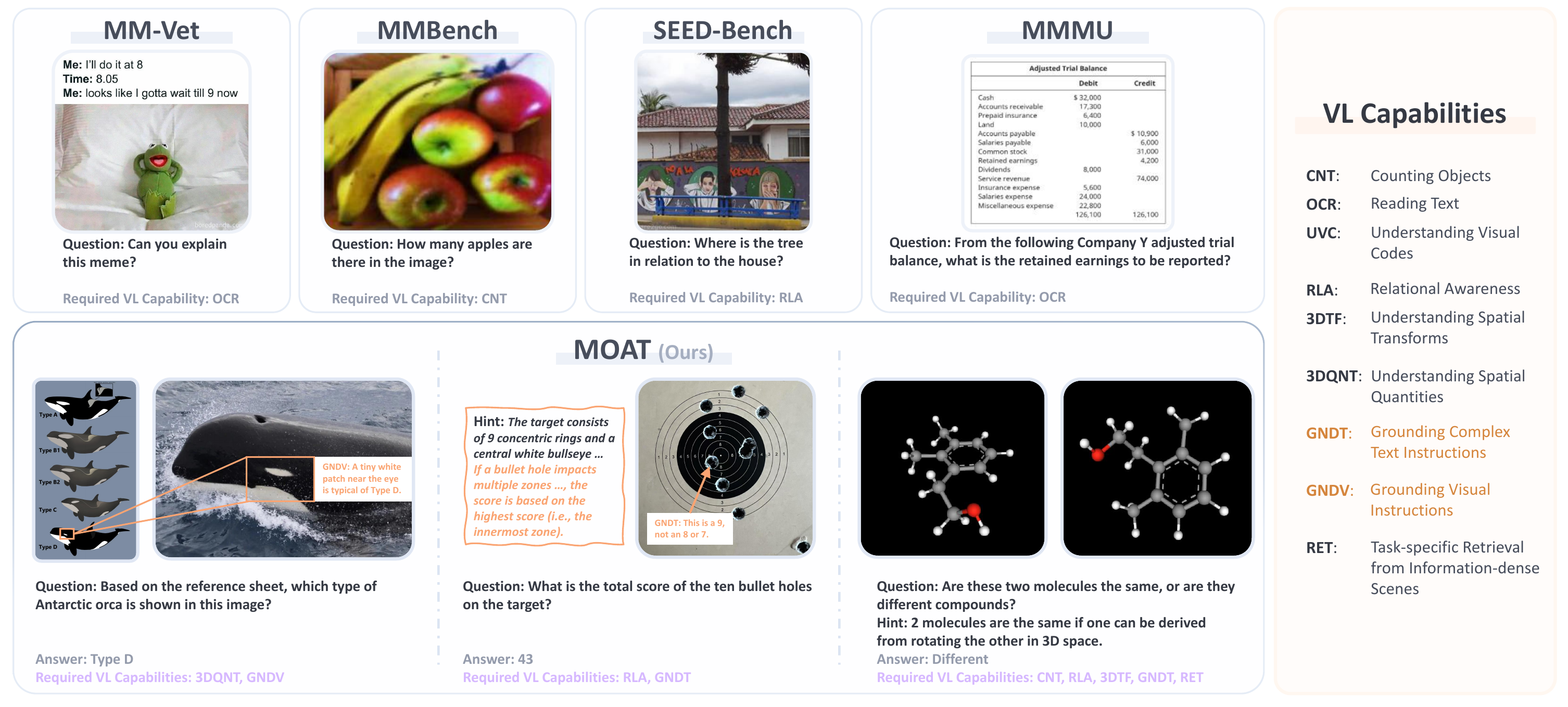}
    \captionof{figure}{Comparison between the tasks in MOAT and existing LMM benchmarks. MOAT tasks are more challenging and better capture the complexity of real-world problems. Specifically, MOAT tasks evaluate LMMs for the ability to \textcolor{color_grounding}{\textbf{ground visual instructions}} (bottom left), \textcolor{color_grounding}{\textbf{ground text instructions}} (center), and \textcolor{color_integrate}{\textbf{integrate a combination of several VL capabilities}} (bottom right). In addition, MOAT tasks are close-ended and complete with hints that provide necessary external knowledge, allowing for fair evaluation of VL capabilities. Please refer to \cref{fig:taxonomy-comparison} and \cref{subsec:taxonomy} for the detailed definition of VL capabilities.}
    \label{fig:figure-1}
    \end{center}
}]

\begin{abstract}
Large multimodal models (LMMs) have demonstrated significant potential as generalists in vision-language (VL) tasks. However, adoption of LMMs in real-world tasks is hindered by their poor performance in tasks that require a combination of VL capabilities, as well as in tasks that involve the grounding of complex text or visual instructions. To thoroughly investigate this gap and its underlying causes, we propose MOAT, a diverse benchmark with 1005 complex real-world vision questions that are straightforward for humans but challenging for LMMs. Specifically, the tasks in MOAT require LMMs to engage in generalist problem solving by integrating VL capabilities such as reading text, counting, understanding spatial relations, grounding textual and visual instructions, etc. All these abilities fit into a taxonomy proposed by us that contains 9 VL capabilities, enabling MOAT to provide a fine-grained view of LMMs' strengths and weaknesses. Besides, MOAT is the first benchmark to explicitly evaluate LMMs' ability to ground complex text and visual instructions, which is essential for many real-world applications. We evaluated 17 proprietary and open source LMMs, finding that the best performing LMM (Gemini 2.5 Pro) achieved only 44\% accuracy, far below what would be acceptable in real-world applications. To guide future model development, we analyze common trends in our results and discuss the underlying causes of poor performance, focusing on the impact of text-centric reasoning, which VL capabilities form bottlenecks in complex tasks, and the potential harmful effects of tiling. Code and data are available at the \href{https://cambrian-yzt.github.io/MOAT/}{project page}.
\end{abstract}    
\section{Introduction}
\label{sec:intro}
% LMMs show great promise (merge with Paragraph 2 if more space is needed)

Vision is the most highly developed sensory modality in humans and forms the basis of how we perceive and understand the world around us \cite{kandel2000principles}. We rely on visual input to solve complex problems in the physical world, including but not limited to navigation, social interaction, and professional tasks (\eg reading a financial chart, CT imagery, or a figure in an academic paper). Recent developments in large multimodal models (LMMs) equip large language models (LLMs) with vision capabilities by adding a vision encoder into the model architecture. These LMMs, such as state-of-the-art examples like GPT 5 and Gemini 2.5 Pro, have shown promise in solving complex vision-language (VL) tasks, such as reading charts, using maps, explaining memes, following instructions, \etc. 

% What sort of benchmark do we need
However, state-of-the-art LMMs still struggle in complex real-world tasks \cite{zhang2024mme, tamarapalli2025countqa, zhang2025lmms}, limiting practical application. This calls for benchmarks that evaluate generalist visual problem solving, in addition to specialist benchmarks evaluating a single capability. Specifically, these general LMM benchmarks should focus on LMMs' ability to (1) effectively combine several VL capabilities (\eg recognition, counting, spatial understanding) at once \cite{yu2024mmvet} and (2) accurately ground detailed instructions in scenes \cite{zhang2024llava}, both of which are essential in practical applications.

\begin{figure*}[htbp]
    \centering
    \includegraphics[width=0.9\textwidth]{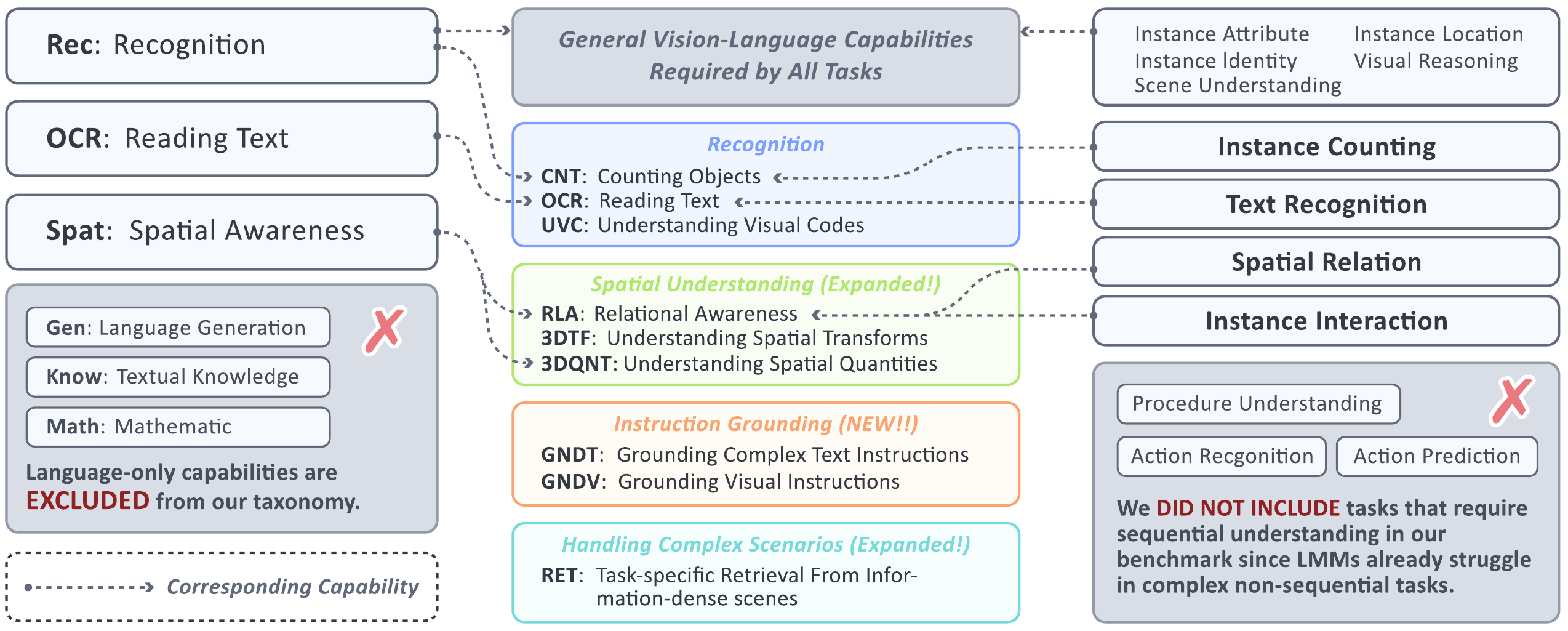}
    \caption{Modifying and expanding upon the capability taxonomy of existing general LMM benchmarks, MOAT's taxonomy focuses on complex tasks and instruction grounding. The emphasis on \textcolor{color_grounding}{\textbf{instruction grounding}} enables MOAT to measure LMMs' capability to make sense of text and visual instructions in actual images, which is neglected by existing benchmarks. Furthermore, we divided \textcolor{color_spatial}{\textbf{spatial understanding}} into fine-grained components. We also systematically define the capability to understand visual codes, and the capability to \textcolor{color_complex_scene}{\textbf{handle complex and noisy scenes}}, which are tested in previous benchmarks but not clearly defined as individual capabilities. To focus on VL capabilities, we purposefully exclude language-only capabilities, which are known to skew results of VL evaluation \cite{xie2025vlmsreadyautonomousdriving, huang2024mmevalpro, liu2024seeing}. Finally, some of the capabilities defined by previous benchmarks are required by all MOAT tasks, and were not included in our taxonomy.}
    \label{fig:taxonomy-comparison}
\end{figure*}

% Why are current benchmarks not enough?
Although MM-Vet \cite{yu2024mmvet}, MMBench \cite{mmbench_eccv2024} and SEED-Bench \cite{seedbench_cvpr2024} have made progress in evaluating VL capabilities and their integration in generalist VQA, they (as shown in \cref{fig:taxonomy-comparison}) often do not cover the full complexity of real-world vision tasks, especially regarding instruction grounding. Moreover, the skill taxonomies in these benchmarks are not enough for fine-grained performance analysis, creating a pressing need for LMM benchmarks that enable fine-grained evaluation of VL capabilities in challenging real-world vision tasks. Finally, many existing benchmarks place a heavy demand on the model's textual knowledge base, which interferes with the evaluation of VL capabilities.

% What sort of new stuff does MOAT bring
To this end, we introduce MOAT~\footnote{MOAT stands for \textbf{M}ultimodal model \textbf{O}f \textbf{A}ll \textbf{T}rades. We believe the capabilities defined in this paper form the \textit{moat} keeping LMMs out of many real-world applications.}, a diverse benchmark with 1005 challenging real-world questions accompanied by a taxonomy that includes 9 VL capabilities key to the practical application of LMMs (see \cref{subsec:taxonomy} for details). This allows MOAT to provide granular insight on how LMMs perform with regard to each VL capability. To reflect the complexity of real-world applications, the questions in MOAT are designed to require the integration of up to 6 VL capabilities. In addition, a significant portion of the tasks in MOAT also requires the model to ground detailed instructions given as text or image (see \cref{fig:figure-1}), an area underexplored by existing benchmarks. Finally, MOAT enables fair VL-only comparison between LMMs by providing all necessary domain knowledge as hints in the question itself. This singles out VL capabilities and levels the playing field regarding factors like textual knowledge base \cite{yu2024mmvet, yue2023mmmu}.

% Our findings
We evaluated 17 LMMs on MOAT. Our key findings are:
\begin{itemize}
    \item MOAT poses significant challenges for LMMs. The best performing model, Gemini 2.5 Pro, has an accuracy of only 44\%. In contrast, humans achieve over 80\% accuracy.
    \item LMMs perform very poorly in counting, relational awareness, and the grounding of text and visual instructions, weaknesses that should be addressed by future models.
    \item Test-time scaling \cite{chen2024expanding} through chain-of-thought (CoT) reasoning brings mixed results in complex VL tasks, failing to consistently improve overall accuracy. Our results show that text-centric CoT reasoning improves context-dependent VL capabilities, while hindering capabilities dependent on visual and spatial understanding.
    \item We use simplified questions that reduce the demand on certain VL capabilities to identify the bottleneck for LMMs. We demonstrate that the bottleneck capabilities of different LMMs are different.
    \item We observe that avoiding tiling by resizing images to the size of one vision encoder tile significantly improves some LMMs' ability to count objects. This calls into question the negative impact on some VL capabilities of certain architectural choices.
\end{itemize}

\section{Related Work}
\label{sec:related-work}

\subsection{General LMM Benchmarks}

% With the rapid rise of LMMs, numerous benchmarks have emerged for evaluating the capabilities of LMMs.
% The progress of LMM benchmarks is closely tied to the development of LMMs.
% In the pre-LLM and LMM era, the multimodal capabilities of deep models were limited, and VL benchmarks (e.g., VQA~\cite{vqa_iccv2015}, VQA v2~\cite{vqav2_cvpr2017} and OK-VQA~\cite{okvqa_cvpr2019}) evaluated basic perception for images, such as the ability to recognize objects or colors.

% With the emergence and development of LMMs, a series of LMM benchmarks have been proposed to evaluate visual perception, knowledge, and reasoning abilities.

% To accurately and comprehensively evaluate LMMs, these benchmarks typically cover a variety of scenarios or VL capabilities, making them as comprehensive and challenging as possible.

Pre-LMM VL benchmarks (\eg, VQA~\cite{vqa_iccv2015}, VQA v2~\cite{vqav2_cvpr2017}, OK-VQA~\cite{okvqa_cvpr2019}), which evaluate basic perception (\eg object recognition, attribute recognition, \etc) in general scenes, have been largely saturated by LMMs. Therefore, more challenging benchmarks have been designed for LMMs. One line of such benchmarks are crafted to cover a wide variety of scenarios and VL capabilities, and strives to enable comprehensive evaluation of LMMs. 

For example, MMMU~\cite{yue2023mmmu} is a multidisciplinary benchmark evaluating college or high school knowledge and reasoning abilities. MMMU-Pro~\cite{MMMU-Pro} further assesses the ability to read questions from images, while Uni-MMMU~\cite{Uni-MMMU} evaluates image understanding and generation simultaneously. General-Bench~\cite{General-Bench} evaluates the synergy between text, vision and audio modalities. MMBench~\cite{mmbench_eccv2024}, SEED-Bench~\cite{seedbench_cvpr2024, li2023seedbench2benchmarkingmultimodallarge, li2024seedbench2plusbenchmarkingmultimodallarge}, MMT-Bench~\cite{mmt_bench_icml2024}, MMStar~\cite{MMStar}, LVLM-EHub~\cite{Lvlm-ehub} and MEGA-BENCH~\cite{MEGA-Bench} each proposes a taxonomy of VL capabilities or scenarios, with each question corresponding to a single capability. These taxonomies often strive to be as comprehensive as possible, assessing LMMs across a wide range of different scenarios. MM-Vet~\cite{yu2024mmvet, yu2024mmvetv2challengingbenchmark} moves a step further, with each question potentially requiring the combination of multiple capabilities, making it more representative of complex real-world scenarios. Finally, MME-RealWorld~\cite{MME-RealWorld} and WildVision~\cite{lu2024wildvision} focus on questions derived from real-world environments. However, LMMs evolve rapidly, and have begun to saturate existing benchmarks, with SOTA LMMs approaching or even surpassing the performance of human experts \cite{zhu2025internvl3, qwen2.5-VL, huang-etal-2025-mmevalpro, wang2025traceable, hong2025glm, zhou2025perception, wang10capabilities}.

\subsection{Specialized LMM Benchmarks}

% To address the saturation of benchmarks results and to more accurately evaluate LMMs’ capabilities in specific areas, many benchmarks targeting specialized skills or scenarios have been introduced.
Another family of benchmarks evaluates LMMs in highly specific areas, trading breadth for depth. MC-Bench~\cite{xu2025mc} emphasizes evaluation in multi-image scenarios. SRBench~\cite{srbench}, VSI-Bench~\cite{yang2024thinking}, 3DSRBench~\cite{ma20253dsrbench} and MMSI-Bench~\cite{yang2025mmsi} evaluate spatial intelligence. DSI-Bench~\cite{zhang2025dsi} and VLM4D~\cite{zhou2025vlm4d} evaluate spatiotemporal intelligence.
PhysBench~\cite{PhysBench} evaluates LMMs’ perception and understanding of the physical world.
GEOBench-VLM~\cite{danish2025geobench} addresses geospatial tasks. MotionBench~\cite{hong2025motionbench} emphasizes fine-grained motion comprehension.
NaturalBench~\cite{li2024naturalbench} and MERLIM~\cite{merlim} argue that existing benchmarks struggle to expose LMMs’ hidden hallucinations (\ie cases where models produce correct answers but lack genuine visual grounding) and propose a set of questions specifically designed to identify them. In addition, several works examine LMM performance in specialized scenarios, such as EmbodiedBench~\cite{yang2025embodiedbench} and VLABench~\cite{zhang2025vlabench} for embodied AI, and RSIEval~\cite{hu2025rsgpt} for remote sensing. VL-RewardBench~\cite{li2025vl-RewardBench} and VLRMBench~\cite{ruan2025vlrmbench} are designed to evaluate the capabilities of multimodal reward models. These benchmarks offer insight on LMM performance in specific areas, and complement the more generalist benchmarks.

\subsection{What Distinguishes MOAT}

% 1. 难度高，贴近复杂现实场景。特别是有instruction grounding，对现实应用更有意义
% 2. 多样的、复合能力。提供了分类体系、以及对应的细粒度评测、分析、诊断
% 3. 所需知识self-contained，只评测模型在VL上的能力、而去除自身知识的影响，评测更公平

Existing general LMM benchmarks trend towards saturation \cite{huang-etal-2025-mmevalpro, wang2025traceable, zhou2025perception, wang10capabilities}, while specialized benchmarks lack comprehensiveness. In contrast, our proposed benchmark, MOAT, is designed to be both comprehensive and challenging. Similar to MM-Vet~\cite{yu2024mmvet}, MOAT evaluates LMMs' integrated capabilities in diverse scenarios, reflecting the complexity of real-world tasks. A key difference from MM-Vet is our inclusion of \textbf{\textit{instruction grounding}}, a skill essential for many practical applications, in our VL capability taxonomy, as well as the division of \textbf{\textit{spatial understanding}} into finer-grained capabilities. Moreover, we provide fine-grained evaluation and diagnostic analysis for each capability, offering insights for improving future LMMs. The challenging nature of MOAT, coupled with the fine-grained diagnostics afforded by our taxonomy, allows for a more in-depth evaluation while maintaining the breadth of the benchmark. Finally, we design MOAT questions to be self-contained in terms of knowledge. Specifically, each question in MOAT is either solvable with common sense or accompanied by the necessary domain-specific information. This allows MOAT to level the playing field in terms of knowledge base, and instead focus on assessing vision–language capabilities, resulting in a fairer evaluation. See \cref{tab:comparison} for a detailed comparison.

% Our capability taxonomy additionally includes instruction grounding, a skill essential for practical applications, further underscoring the relevance of MOAT to real-world usage.
% Moreover, we provide a fine-grained evaluation and diagnostic analysis for each capability, offering insights for improving future LMMs.
% Finally, we intentionally design MOAT to require only self-contained knowledge.
% Specifically, each question in MOAT either relies solely on common sense knowledge or provides any necessary domain-specific information through extra text or images.
% This allows MOAT to focus on assessing vision–language capabilities while minimizing the influence of expert knowledge with LMMs, resulting in a more fair evaluation.

\newcommand{\goodyes}{\textcolor{green}{\faCheck}}
\newcommand{\badno}{\textcolor{red}{\faTimes}}
\newcommand{\badyes}{\textcolor{red}{\faCheck}}
\newcommand{\goodno}{\textcolor{green}{\faTimes}}
\newcommand{\halfyes}{\textcolor{yellow}{\faCheck\llap{\faTimes}}}

\begin{table*}[hbtp]
\resizebox{\textwidth}{!}{%
    \begin{tabular}{l|c|c|ccc|ccc|cc|c|c}
    \toprule
          & \multirow{2}[4]{*}{\textbf{Taxonomy}} & \multicolumn{1}{c|}{\multirow{2}[4]{*}{\textcolor{color_integrate}{\textbf{\makecell{Integration \\ of Cap.}}}}} & \multicolumn{3}{c|}{\textcolor{color_recognition}{\textbf{Recognition}}} & \multicolumn{3}{c|}{\textcolor{color_spatial}{\textbf{Spatial Understanding}}} & \multicolumn{2}{c|}{\textcolor{color_grounding}{\textbf{Instruction Grounding}}} & \multicolumn{1}{c|}{\textcolor{color_complex_scene}{\textbf{Complex Scenes}}} & \multirow{2}[4]{*}{\textbf{\makecell{Text-only \\ Capabilities}}} \\
\cmidrule{4-12}          &       &       & \textcolor{color_recognition}{\textbf{CNT}}   & \textcolor{color_recognition}{\textbf{OCR}}   & \textcolor{color_recognition}{\textbf{UVC}}   & \textcolor{color_spatial}{\textbf{RLA}}   & \textcolor{color_spatial}{\textbf{3DTF}}  & \textcolor{color_spatial}{\textbf{3DQNT}} & \textcolor{color_grounding}{\textbf{GNDT}}  & \textcolor{color_grounding}{\textbf{GNDV}}  & \textcolor{color_complex_scene}{\textbf{RET}} &  \\
    \midrule
    MMMU~\cite{yue2023mmmu} & \badno & - & - & - & - & - & - & - & - & - & - & - \\
    MM-Vet v2~\cite{yu2024mmvetv2challengingbenchmark} & \goodyes & \goodyes & \goodyes & \goodyes & \badno & \goodyes & \badno & \goodyes & \badno & \badno & \badno & \badyes \\
    MMBench~\cite{mmbench_eccv2024} & \goodyes & \badno & \goodyes & \goodyes & \halfyes & \goodyes & \badno & \badno & \badno & \badno & \badno & \badyes \\
    SEED-Bench-2-Plus~\cite{li2024seedbench2plusbenchmarkingmultimodallarge} & \goodyes & \badno & \goodyes & \goodyes & \badno & \goodyes & \badno & \badno & \badno & \badno & \badno & \badyes \\
    MMT-Bench~\cite{mmt_bench_icml2024} & \goodyes & \badno & \goodyes & \goodyes & \halfyes & \goodyes & \badno & \badno & \badno & \halfyes & \goodyes & \badyes \\
    MEGA-Bench~\cite{MEGA-Bench} & \goodyes & \badno & \goodyes & \goodyes & \halfyes & \goodyes & \badno & \badno & \badno & \badno & \goodyes & \badyes \\
    VSI-Bench~\cite{yang2024thinking} & \goodyes & \badno & \goodyes & \badno & \badno & \goodyes & \badno & \goodyes & \badno & \badno & \badno & \badyes \\
    General-Bench~\cite{General-Bench} & \goodyes & \badno & \goodyes & \goodyes & \halfyes & \goodyes & \badno & \goodyes & \badno & \badno & \goodyes & \badyes \\
    MMStar~\cite{MMStar} & \goodyes & \badno & \goodyes & \goodyes & \halfyes & \goodyes & \badno & \badno & \badno & \badno & \badno & \badyes \\
    LVLM-EHub~\cite{Lvlm-ehub} & \goodyes & \badno & \goodyes & \goodyes & \halfyes & \badno & \badno & \badno & \badno & \badno & \goodyes & \badyes \\
    WildVision~\cite{lu2024wildvision} & \goodyes & \badno & \badno & \goodyes & \halfyes & \goodyes & \badno & \badno & \badno & \badno & \goodyes & \badyes \\ \midrule
    MOAT (Ours) & \goodyes & \goodyes & \goodyes & \goodyes & \goodyes & \goodyes & \goodyes & \goodyes & \goodyes & \goodyes & \goodyes & \goodno \\ \bottomrule
    \end{tabular}%
}
\caption{How our taxonomy compare to previous benchmarks. While some taxonomies of VL capabilities are provided by these benchmarks, they are far from comprehensive, and usually mixed with text-modality-only capabilities. Only MOAT and MM-Vet series~\cite{yu2024mmvet, yu2024mmvetv2challengingbenchmark} examine the integration of multiple capabilities (Integration of Cap.), which makes the questions more complex and closer to real-world problems. Furthermore, few existing benchmarks consider the ability to ground complex textual and visual instructions (GNDT \& GNDV). MMT-Bench~\cite{mmt_bench_icml2024}, the only exception, includes a meta-task called Cross-Image Matching, which partially overlaps with GNDV.}
\label{tab:comparison}
\end{table*}

\section{MOAT}
\label{sec:benchmark}
Three characteristics differentiate MOAT from existing LMM benchmarks:
(1) MOAT consists of questions that require LMMs to integrate multiple VL capabilities simultaneously, which makes MOAT challenging and closer to real-world problems. (2) MOAT evaluates LMMs for the capability to ground visual instructions and complex text instructions, which is neglected by previous benchmarks but essential for many real-world applications. (3) The questions in MOAT are designed to be close-ended and solvable with the knowledge and hints included in the question itself. This enables a fair comparison between LMMs.

\subsection{Taxonomy of Vision-Language Capabilities}
\label{subsec:taxonomy}

Expanding upon the capabilities mentioned in existing benchmarks such as MM-Vet \cite{yu2024mmvet} and SEED-Bench \cite{seedbench_cvpr2024} (see \cref{fig:taxonomy-comparison}), we define 9 VL capabilities. Specifically, counting and OCR are challenging categories carried over from existing taxonomies. In addition to these, we added the capability of understanding visual codes, a prevalent type of information in real world. Meanwhile, we subdivided the \textit{spatial relation} category of existing benchmarks into finer-grained components. Finally, we added capabilities regarding instruction grounding and handling complex scenes. These are crucial to real-world applications and are under-represented by previous benchmarks. 

We did not include the \textit{capabilities required by all MOAT tasks}, such as object recognition and attribute recognition, in our taxonomy. These capabilities are well-studied and no longer pose challenges for LMMs \cite{ross2025s, seedbench_cvpr2024, zhou2025human}. We also purposefully excluded \textit{text-only capabilities} (\eg language generation and math), which are known to skew the results of VL capability evaluation \cite{xie2025vlmsreadyautonomousdriving}.

\vspace{-0.4cm}
\paragraph{\tagrec{Recognition}}
\begin{itemize}
    \item \textbf{Counting (\tagrec{CNT})}: the ability to accurately count objects in an image.
    \item \textbf{Optical Character Recognition (\tagrec{OCR})}: the ability to read text in an image.
    \item \textbf{Understanding Visual Codes (\tagrec{UVC})}: the ability to understand visual codes designed for humans, \eg the legend of a figure, signs, icons, \etc.
\end{itemize}
\vspace{-0.5cm}

\paragraph{\tagspatial{Spatial Understanding}}
\begin{itemize}
    \item \textbf{Awareness of Spatial Relation (\tagspatial{RLA})}: the ability to recognize the spatial relation between objects. This also includes the ability to understand how objects are physically connected.
    \item \textbf{Understanding Spatial Transforms (\tagspatial{3DTF})}: the ability to understand 3D spatial transforms (\eg rotation, reflection) and their effects on the semantics of objects. For example, rotation changes the direction of an arrow, but does not change the chemical properties of a molecule.
    \item \textbf{Understanding Spatial Quantities (\tagspatial{3DQNT})}: the ability to estimate and compare spatial quantities (\eg length, angle, area, volume, \etc) in an image.
\end{itemize}
\vspace{-0.5cm}

\paragraph{\taggnd{Instruction Grounding}}
\begin{itemize}
    \item \textbf{Grounding of Text Instructions (\taggnd{GNDT})}: the ability to make sense of and follow complex text instructions (\eg the rules for calculating the score of an archery target) when solving VL problems. This ability is essential in the application of LMMs in-the-wild.
    \item \textbf{Grounding of Visual Instructions (\taggnd{GNDV})}: the ability to follow image-based instructions (\eg a Lego instruction manual). \taggnd{GNDV} is especially relevant in scenarios where visual instructions are more convenient.
\end{itemize}
\vspace{-0.5cm}

\paragraph{\tagcomplex{Handling Complex Scenarios}}
\begin{itemize}
    \item \textbf{Task-Specific Retrieval from Dense Scenes (\tagcomplex{RET})}: the ability to retrieve task-specific clues from images with high information density. For example, signs in a complex train station can point to dozens of lines and exits, and \tagcomplex{RET} is required to find the relevant sign.
\end{itemize}

\vspace{0.2cm}
\subsection{Building MOAT}
\label{subsec:data-collection}

MOAT consists of more than 1000 images, some of which we took ourselves while others were sourced from the web. For web images, we strictly followed copyright laws and licensing. Based on these images, we crafted 1005 questions, with each question requiring the model to understand one or several images. To facilitate fair and objective evaluation of the 9 VL capabilities, we crafted the questions to have brief and unambiguous answers to minimize the influence of language generation style and simplify the evaluation process. Furthermore, we included necessary external knowledge (\eg the orca classification reference sheet and the text hints in \cref{fig:figure-1}) in questions requiring knowledge beyond commonsense to create a leveled playing field (in terms of VL capabilities) for different models. 

We quality-checked each question to ensure that (1) it has an unambiguous answer, (2) the external knowledge provided, if any, is adequate for a human without prior knowledge of related fields, and (3) the question is straightforward for humans. After the questions were designed, 4 researchers labeled all 1005 questions independently. The resulting conflicts were resolved collaboratively through discussion. The breakdown of individual VL capabilities and capability combinations required is shown in \cref{fig:dataset-composition}. See Appendix for example questions and more details on the dataset.

\begin{figure}[htbp]
    \centering
    \includegraphics[width=0.45\textwidth]{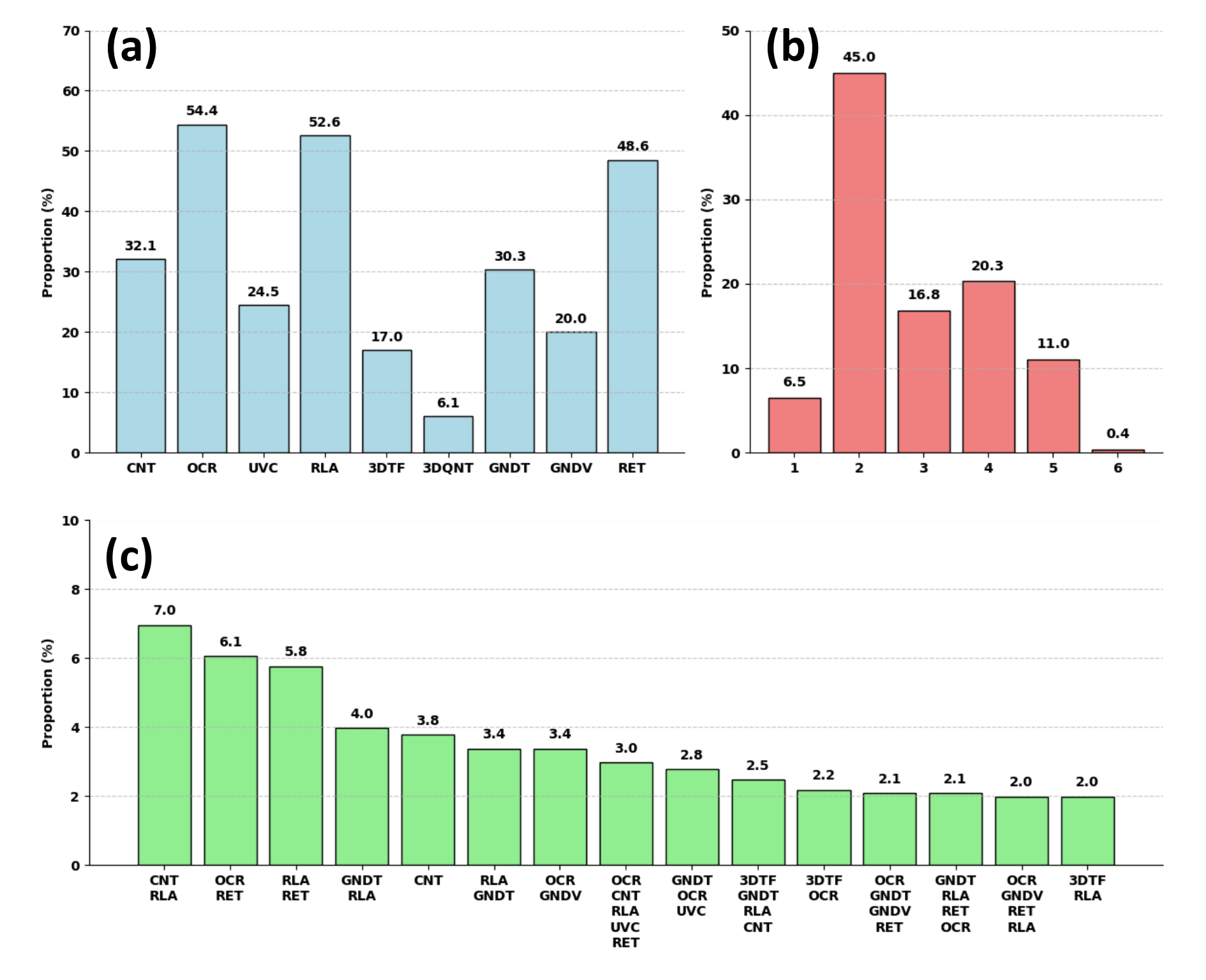}
    \caption{The proportion of questions requiring each VL capability is shown in (a), while the distribution of the number of capabilities needed to solve problems is shown in (b), demonstrating the complexity of MOAT tasks. The 15 (out of 82) most common capability combinations is shown in (c).}
    \label{fig:dataset-composition}
\end{figure}
\section{Experiments}
\label{sec:experiments}
\subsection{Experiment Settings}
\label{subsec:exp-settings}

\begin{table*}[htbp]
\centering
\small

\begin{tabular}{lcccccccccc}
\hline
Model                                         & \tagrec{CNT}         & \tagrec{OCR}         & \tagrec{UVC}         & \tagspatial{RLA}     & \tagspatial{3DTF}    & \tagspatial{3DQNT}   & \taggnd{GNDT}        & \taggnd{GNDV}        & \tagcomplex{RET}     & \textbf{Overall}              \\ \hline
\textit{Human*}                               & \textit{92.66}       & \textit{82.99}       & \textit{72.92}       & \textit{78.57}       & \textit{78.86}       & \textit{77.78}       & \textit{82.32}       & \textit{70.83}       & \textit{84.62}       & \textit{82.72$\pm$9.11}       \\ \hline
Gemini 2.5 Pro                                & {\ul \textbf{39.32}} & {\ul \textbf{46.13}} & 47.15                & 40.01                & 43.47                & {\ul \textbf{45.90}} & 31.69                & {\ul \textbf{49.42}} & {\ul \textit{46.24}} & {\ul \textbf{44.01$\pm$1.79}} \\
GPT 5                                         & {\ul \textit{39.22}} & {\ul \textit{44.67}} & {\ul \textit{48.10}} & {\ul \textbf{42.53}} & {\ul \textit{47.17}} & {\ul \textbf{45.90}} & {\ul \textit{34.21}} & {\ul \textit{47.26}} & {\ul \textbf{46.45}} & {\ul \textit{43.88$\pm$0.92}} \\
GPT 5 Mini                                    & 35.40                & 43.69                & {\ul \textbf{49.86}} & {\ul \textit{40.83}} & 42.30                & 27.87                & {\ul \textbf{34.54}} & 43.95                & 45.01                & 41.63$\pm$0.77                \\
GPT 4.1                                       & 36.43                & 36.87                & 40.92                & 39.45                & {\ul \textbf{48.34}} & 27.32                & 31.37                & 37.65                & 38.59                & 38.28$\pm$0.81                \\
Gemini 2.5 Flash                              & 36.33                & 37.72                & 36.99                & 34.09                & 45.03                & 35.52                & 26.01                & 38.47                & 37.84                & 37.55$\pm$1.14                \\
GPT 4.1 Mini                                  & 34.47                & 36.38                & 37.40                & 37.68                & 44.83                & 27.32                & 28.31                & 37.48                & 37.23                & 36.65$\pm$0.09                \\
Claude Sonnet 4.5                             & 30.96                & 37.42                & 38.21                & 36.74                & 44.44                & 25.14                & 28.52                & 38.31                & 38.32                & 35.92$\pm$0.61                \\
Claude Opus 4                                 & 31.79                & 34.92                & 34.42                & 35.60                & 44.05                & 22.95                & 27.21                & 38.97                & 35.72                & 34.89$\pm$0.87                \\
GPT 4o                                        & 34.78                & 31.87                & 36.31                & 33.96                & 41.91                & 30.60                & 31.58                & 30.18                & 34.97                & 34.06$\pm$0.65                \\
Doubao Seed 1.6                               & 27.04                & 34.67                & 34.82                & 32.01                & 36.06                & 24.59                & 32.13                & 37.31                & 33.13                & 33.23$\pm$0.49                \\
\cellcolor[HTML]{DAE8FC}GLM 4.5v              & 27.66                & 33.70                & 38.48                & 31.06                & 37.62                & 32.24                & 26.89                & 32.67                & 32.31                & 32.77$\pm$0.63                \\
\cellcolor[HTML]{DAE8FC}Qwen3 30B A3B Think   & 30.65                & 31.26                & 34.96                & 33.27                & 37.43                & 27.87                & 26.89                & 30.02                & 35.93                & 32.11$\pm$0.40                \\
Claude Haiku 4.5                              & 30.34                & 30.29                & 36.18                & 32.51                & 36.45                & 21.86                & 33.44                & 31.84                & 32.92                & 31.87$\pm$0.20                \\
\cellcolor[HTML]{DAE8FC}Qwen3 235B A22B Think & 30.75                & 29.19                & 31.57                & 33.02                & 36.65                & 27.87                & 25.79                & 28.19                & 34.29                & 31.21$\pm$0.48                \\
Gemini 2.5 Flash Lite                         & 28.17                & 26.93                & 26.29                & 26.78                & 36.65                & 24.59                & 20.44                & 28.19                & 29.92                & 28.26$\pm$1.95                \\
GPT 5 Nano                                    & 24.97                & 26.45                & 35.50                & 30.06                & 37.62                & 21.86                & 28.42                & 23.38                & 26.98                & 28.19$\pm$1.12                \\
GPT 4.1 Nano                                  & 19.50                & 20.96                & 26.83                & 27.28                & 38.01                & 18.58                & 21.64                & 20.40                & 23.77                & 23.18$\pm$0.41                \\ \hline
\textit{Random Guessing}                      & -                    & -                    & -                    & -                    & -                    & -                    & -                    & -                    & -                    & \textit{14.41}                \\ \hline
\end{tabular}

\caption{Results of the main experiment. The top performing model is {\ul \textbf{bolded}}, while the runner up is {\ul \textit{italicized}}. A blue background denotes open-source models. The performance in each VL capability is measured by the model's accuracy in all questions requiring that capability. The overall accuracy is not the average per-capability accuracy because the capabilities are unevenly spread. The random guess baseline is obtained by randomly guessing multiple-choice questions and giving up altogether on fill-in-the-blank questions. *Human performance measured using a 189-question subset of MOAT and serves only as a rough estimation demonstrating the large human-LMM gap.}
\label{tab:main}
\end{table*}

We evaluated popular proprietary and open source LMMs on MOAT. We ran the evaluation three times for each model to account for the randomness of LMM output, iterating through all 1005 questions in each run. For multiple choice questions, the choices were randomly shuffled each time to obtain objective results. All evaluations were zero-shot. We adopted the standard LLM-as-a-judge approach \cite{yu2024mmvet, lu2024wildvision, huang2024empirical} and used GPT 4.1 to compare the output with the ground truth, resulting in a binary classification of \textit{right} or \textit{wrong}. We manually inspected the logs of all 1005 questions in 3 runs and disagreed with the LLM judge only 5 out of 3015 times, confirming the overall reliability of the evaluation procedure. We set the temperature to $0$ for all models except the GPT 5 family, which only accepted $1.0$ as the temperature value. All models are evaluated using a system prompt that contains a simple chain-of-thought (CoT) instruction (see Appendix for prompts), where the model is asked to analyze the problem first before answering the question. 

In addition, we recruited 3 graduate students to complete a 189-question subset of MOAT, which has a distribution of VL capabilities similar to the full benchmark, to provide a rough estimation of human performance.

\subsection{Main Results and Analysis}
\label{subsec:exp-main-results}
We report the main experiment results in \cref{tab:main}, including overall accuracy and performance in individual VL capabilities. We choose accuracy as the main metric because all questions are close-ended and were graded as either right or wrong. We draw the following conclusions from the results.

\textbf{MOAT is far from saturation.} Benchmark saturation, where the rapid improvement in LMM performance renders existing benchmarks obsolete, is a constant challenge in LMM evaluation. State-of-the-art models like GPT 5 and Gemini 2.5 Pro have already reached human-level performance on benchmarks like MMMU \cite{comanici2025gemini}. However, MOAT is still far from saturation, with LMM performance capped at 44\% (vs 83\% human performance). Moreover, comparing the performance of three consecutive generations of GPT models (4o, 4.1 and 5), we see only modest improvements on MOAT at around 5 percentage points per generation. Therefore, we are optimistic that MOAT will remain challenging for next-gen LMMs and stay relevant for years to come.

\textbf{Key capabilities remain undeveloped.} In our experiments, \textbf{CNT}, \textbf{RLA} and \textbf{GNDT} saw consistently poor performance across all models. In addition, apart from the top 3 performers, LMMs also struggled to understand visual codes designed for humans (\textbf{UVC}) and spatial quantities (\textbf{3DQNT}). \textbf{CNT}, \textbf{RLA}, and \textbf{3DQNT} are closely related to the understanding of 3D space and the objects within, and the poor result is consistent with previous studies that show LMMs' lack of spatial awareness \cite{yang2024thinking}. Meanwhile, \textbf{GNDT}, and \textbf{UCV} generally involve intense visual reasoning, as they require the model to make sense of abstract input (\eg instructions and charts) in an image. The inability of existing LMMs to understand 3D space and reason with visual input indicates that there is still a long way to go before LMMs can compete with humans in complex real-world VL tasks.

\begin{figure}[htbp]
    \centering
    \includegraphics[width=0.45\textwidth]{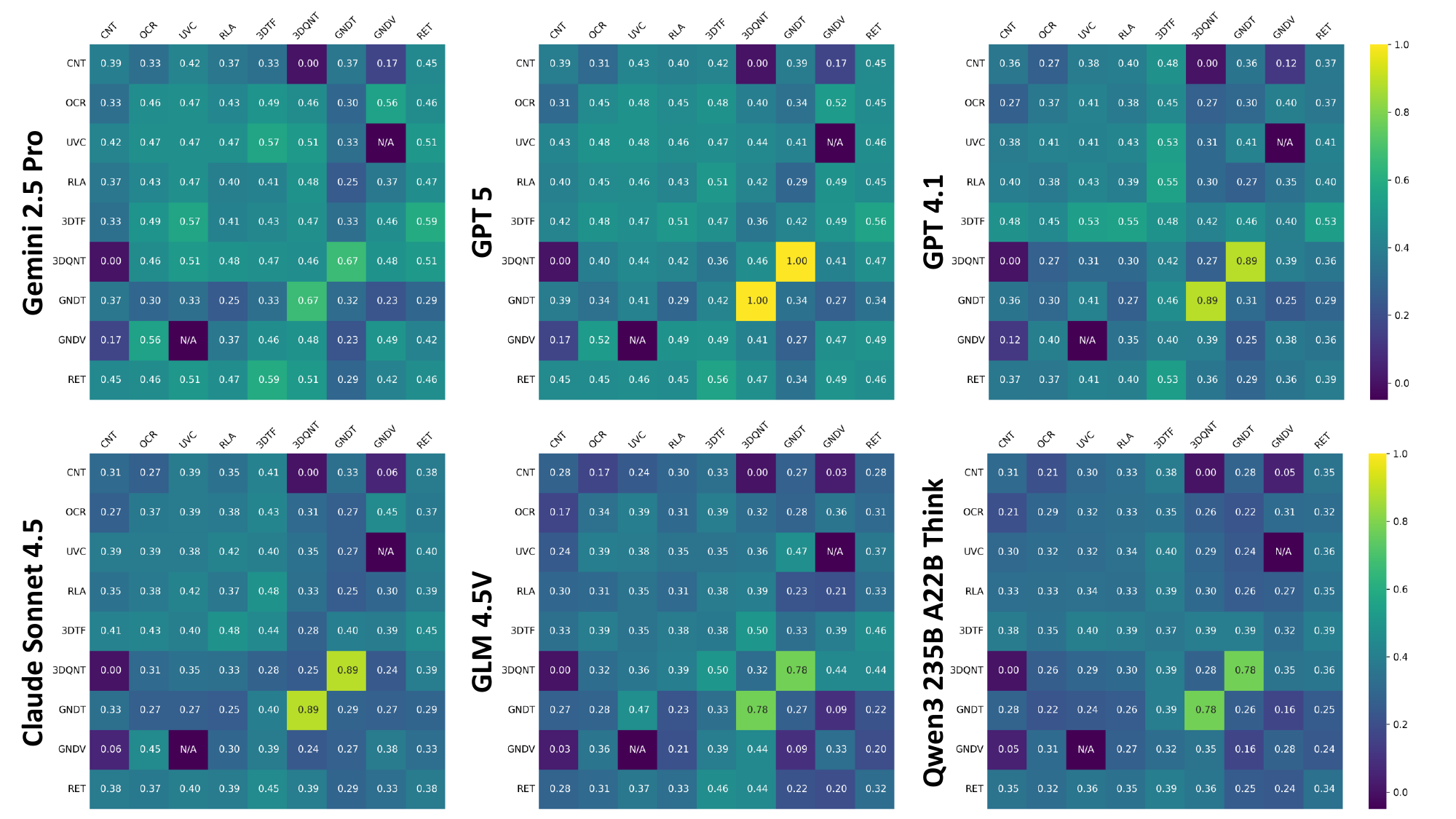}
    \caption{Inter-capability interaction in 6 models. Each cell in the heatmaps represents the accuracy of the model on questions requiring the combination of 2 VL capabilities.}
    \label{fig:heatmap}
\end{figure}

\textbf{How do VL capabilities interact with each other?} MOAT is designed to investigate the interactions between VL capabilities in complex real-world tasks. Here, we visualize these interactions as heatmaps (\cref{fig:heatmap}) to gain more insight on inter-capability correlation. An interesting observation is that, while larger, more-advanced models are better at tasks requiring the integration of recognition and spatial capabilities, they struggle to combine instruction grounding with either recognition or spatial understanding.

\textbf{Are larger models necessarily better?} Scaling law has been the driving force in LLM and LMM development. However, the results on MOAT show that scaling up alone is not enough. For the Claude and Qwen3 model families, the larger models (Opus and Qwen3 235B) performed worse than their smaller counterparts (Sonnet and Qwen3 30B). Moreover, GPT 5 and GPT 4.1 performed only marginally better than their Mini versions. This underscores the limitations of simply scaling up model size in complex VL tasks.

\subsection{Thinking In Text Is Not Enough}
\label{subsubsec:exp-main-thinking}

\begin{table*}[htbp]
\footnotesize
\centering
\setlength{\tabcolsep}{5pt}
\begin{tabular}{lccccccccccc}
\hline
Model                         & \tagrec{CNT}         & \tagrec{OCR}         & \tagrec{UVC}         & \tagspatial{RLA}     & \tagspatial{3DTF}    & \tagspatial{3DQNT}   & \taggnd{GNDT}        & \taggnd{GNDV}        & \tagcomplex{RET}     & \textbf{Overall}     & \begin{tabular}[c]{@{}c@{}}\textbf{Avg.}\\ \textbf{Latency (s)}\end{tabular} \\ \hline
GPT 5 Minimal                 & 32.82                & 33.70                & 35.09                & 35.16                & 40.55                & 24.59                & 26.45                & 34.33                & 36.95                & 34.96                & 9.1                                                        \\
GPT 5 Low                     & 37.05                & 43.39                & 47.43                & 40.20                & {\ul \textbf{47.56}} & 36.61                & 31.58                & 42.95                & 45.08                & 42.42                & 15.5                                                       \\
GPT 5 Medium                  & {\ul \textbf{39.22}} & {\ul \textbf{44.67}} & 48.10                & {\ul \textbf{42.53}} & 47.17                & {\ul \textbf{45.90}} & {\ul \textbf{34.21}} & {\ul \textbf{47.26}} & 46.45                & {\ul \textbf{43.88}} & 33.3                                                       \\
GPT 5 High                    & 38.18                & 43.88                & {\ul \textbf{48.37}} & 40.52                & 45.81                & 41.53                & 32.68                & 43.95                & {\ul \textbf{46.72}} & 43.08                & 71.5                                                       \\ \hline
GPT 5 Mini Minimal            & 34.26                & 35.34                & 37.94                & 36.23                & {\ul \textbf{43.27}} & {\ul \textbf{31.69}} & 28.09                & 38.31                & 38.46                & 36.15                & 13.0                                                       \\
GPT 5 Mini Low                & {\ul \textbf{35.40}} & 42.41                & 46.21                & 39.76                & 41.52                & 31.15                & 34.32                & 43.62                & 42.28                & 40.70                & 13.4                                                       \\
GPT 5 Mini Medium             & 34.06                & 42.29                & 47.56                & 39.57                & 41.52                & 29.51                & {\ul \textbf{36.28}} & 43.12                & 43.44                & 40.53                & 19.0                                                       \\
GPT 5 Mini High               & {\ul \textbf{35.40}} & {\ul \textbf{43.69}} & {\ul \textbf{49.86}} & {\ul \textbf{40.83}} & 42.30                & 27.87                & 34.54                & {\ul \textbf{43.95}} & {\ul \textbf{45.01}} & {\ul \textbf{41.63}} & 54.3                                                       \\ \hline
Gemini 2.5 Flash w/o Thinking & {\ul \textbf{36.33}} & {\ul \textbf{37.72}} & 36.99                & 34.09                & {\ul \textbf{45.03}} & 35.52                & 26.01                & {\ul \textbf{38.47}} & 37.84                & {\ul \textbf{37.55}} & 17.5                                                       \\
Gemini 2.5 Flash Thinking     & 32.61                & 36.08                & {\ul \textbf{37.67}} & {\ul \textbf{36.74}} & 40.16                & {\ul \textbf{39.89}} & {\ul \textbf{27.10}} & 33.50                & {\ul \textbf{40.37}} & 36.48                & 31.9                                                       \\ \hline
\end{tabular}
\caption{Results for 3 models under different reasoning settings. For each model, the reasoning effort increases from top to bottom.}
\label{tab:thinking}
\end{table*}

Thinking models with integrated chain-of-thought (CoT) reasoning have shown great potential in text-centric tasks such as math and coding. Some flagship models (\eg GPT 5 and Gemini 2.5 Pro) even have thinking enabled by default. We explore the effect of test time scaling (\ie facilitating CoT reasoning during inference) on the complex VL tasks in MOAT by evaluating LMMs under different reasoning settings. For GPT 5 and GPT 5 Mini, we modified the \textit{reasoning effort} parameter in API calls, resulting in 4 conditions (minimal, low, medium, and high). Meanwhile, we evaluated Gemini 2.5 Flash with thinking mode both turned on and off, resulting in 2 conditions. We did not evaluate Gemini 2.5 Pro here because a non-thinking version is not available. We report the results in \cref{tab:thinking}.

\textbf{Chain-of-thought is not a magic solution.} The results suggest that chain-of-thought (CoT) reasoning, widely used in text-centric tasks like math and coding, does not bring consistent improvement. For GPT 5 and GPT 5 Mini, the \textit{medium} and \textit{high} settings bring little improvement over the \textit{low} setting in overall performance. For Gemini, the overall accuracy is actually lower with thinking mode on. The shortcomings of text-centric CoT are even more obvious when we consider latency and cost. With the reasoning effort set to high, GPT 5 used thousands of reasoning tokens and more than a minute on average to solve VL tasks that are fairly straightforward for humans (all 3 humans finished 189 questions in under 120 minutes); yet, the result was still quite poor. Therefore, solutions beyond text-centric CoT are needed for real-world VL tasks.

\textbf{Reasoning improves text-dependent and context-dependent capabilities.} Despite mixed results in overall accuracy, MOAT's fine-grained capability taxonomy allows us to discover clear trends regarding individual VL capabilities. For instance, text-based reasoning consistently improves LMM performance in \textbf{UVC} and \textbf{RET}. This result is mostly in line with our expectations, since understanding visual codes (\textbf{UVC}) and retrieving relevant information from noisy scenes (\textbf{RET}) require step-by-step thinking, and the visual information involved can often be clearly described in text. In addition, CoT reasoning's impact on \textbf{OCR} and \textbf{RLA} is also largely positive. We hypothesize that reasoning about the surrounding context could provide clues for \textbf{OCR} in blurry images, and help the LMM identify mistakes in its initial understanding of the relation between objects (\textbf{RLA}). 

\textbf{Reasoning does not improve vision-dominant capabilities.} In contrast, results are mixed for \textbf{CNT}, \textbf{3DQNT}, \textbf{GNDT}, and \textbf{GNDV}, and negative for \textbf{3DTF}. Here, tuning up the reasoning effort does not bring reliable improvement. This is expected for \textbf{CNT}, \textbf{3DTF} and \textbf{3DQNT}, since these rely more on \textit{directly} understanding the image, not step-by-step reasoning. However, the results for \textbf{GNDT} and \textbf{GNDV} are quite surprising, as we assumed that grounding complex instructions benefits from CoT reasoning. We hypothesize that the root cause is LMMs' failure to extract nuanced details from images that align with the instructions. As a result, the bottleneck is perception instead of reasoning, leading to mixed results for thinking models. Specific failure cases can be found in the Appendix.

\textbf{Use thinking models with discretion.} Existing text-centric CoT reasoning does not consistently improve LMM performance in complex vision tasks. Our fine-grained evaluation of how reasoning affects each VL capability sheds light on how to tune the reasoning effort based on the VL capabilities required by the task. Cost and latency stemming from prolonged reasoning should also be considered.

\subsection{Bottleneck Capability Analysis}
\label{subsec:exp-error-analysis}
As shown in \cref{subsec:exp-main-results}, LMMs performed very poorly on counting (\textbf{CNT}), relational awareness (\textbf{RLA}), and grounding text instructions (\textbf{GNDT}). Therefore, we designed additional experiments to probe \textit{why} LMMs fail in these tasks and demonstrate MOAT's potential as a diagnostic tool for LMMs. Specifically, our goal is to identify which capability (or combination of capabilities) forms the bottleneck for each model. Inspired by ablation studies, we created simplified versions of the most challenging questions in MOAT. In the simplified questions, the images were edited to include visual prompts \cite{shtedritski2023does, shao2024visual, vip_llama_cvpr2024} or exclude irrelevant areas (see \cref{fig:ablation-1} and \cref{fig:ablation-2}). This allowed us to precisely reduce difficulty regarding certain VL capabilities in our search for the bottleneck.

\subsubsection{Analyzing \textbf{RLA} and \textbf{GNDT}}
\label{subsubsec:exp-error-analysis-rla-gndt}
\begin{figure}[htbp]
    \centering
    \includegraphics[width=0.95\linewidth]{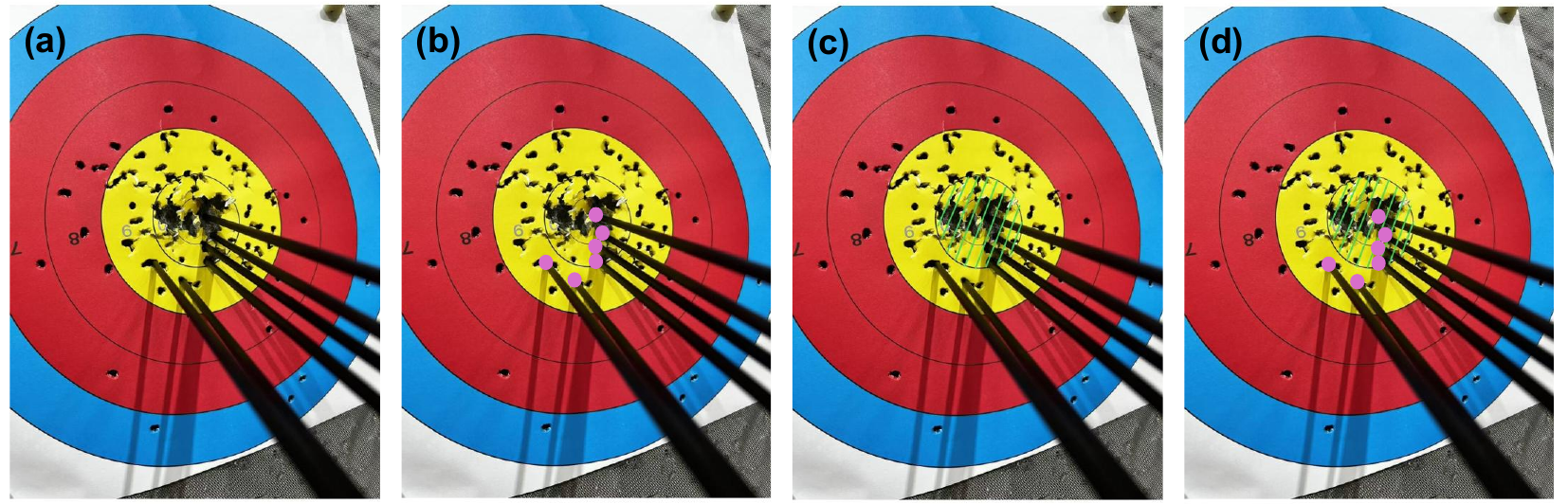}
    \caption{Simplified versions of the task of scoring archery targets: (a) is the original image; (b) simplifies \textbf{RLA} by marking the impact points of arrows; (c) simplifies \textbf{GNDT} by marking the scoring area for 10 on the target; (d) is a combination of both.}
    \label{fig:ablation-1}
\end{figure}

For relational awareness (\textbf{RLA}) and grounding text instructions (\textbf{GNDT}), we consider the task of calculating the score of an archery target (\cref{fig:ablation-1}a). We explicitly marked the arrows' point of impact (\cref{fig:ablation-1}b) to simplify \textbf{RLA}. To help the model ground the text instructions describing scoring areas in the context of each image (\textbf{GNDT}), we marked the scoring area for 10 to differentiate it from the area for 9 (\cref{fig:ablation-1}c), a subtask where LMMs perform badly. Adding both clues results in (\cref{fig:ablation-1}d). We evaluated Gemini 2.5 Flash, GPT 4.1, GPT 4.1 Mini, and Qwen3 235B A22B Think on the simplified tasks. We report the results in \cref{tab:ablation1}.

\begin{table}[htbp]
\centering
\setlength{\tabcolsep}{2pt}
\scriptsize
\begin{tabular}{|l|cc|cc|cc|cc|}
\hline
\multirow{2}{*}{Model} & \multicolumn{2}{c|}{Original}            & \multicolumn{2}{c|}{Simplify \tabspatial{RLA}} & \multicolumn{2}{c|}{Simplify \tabgnd{GNDT}} & \multicolumn{2}{c|}{Simplify \textbf{Both}} \\ \cline{2-9} 
                       & \multicolumn{1}{l|}{Acc}           & MAE & \multicolumn{1}{l|}{Acc}                       & MAE            & \multicolumn{1}{l|}{Acc}                 & MAE               & \multicolumn{1}{l|}{Acc}                 & MAE               \\ \hline
Gemini 2.5 Flash       & \multicolumn{1}{l|}{33.3}          & 1.5 & \multicolumn{1}{l|}{\textbf{41.7}}             & 1.0            & \multicolumn{1}{l|}{25.0}                & 2.4               & \multicolumn{1}{l|}{\textbf{41.7}}       & \textbf{0.9}      \\ \hline
GPT 4.1                & \multicolumn{1}{l|}{30.0}          & 1.5 & \multicolumn{1}{l|}{\textbf{38.3}}             & 1.6            & \multicolumn{1}{l|}{25.0}                & 2.1               & \multicolumn{1}{l|}{35.8}                & \textbf{1.5}      \\ \hline
GPT 4.1 Mini           & \multicolumn{1}{l|}{22.5}          & 2.4 & \multicolumn{1}{l|}{16.7}                      & 3.4            & \multicolumn{1}{l|}{25.8}                & 2.4               & \multicolumn{1}{l|}{\textbf{27.5}}       & \textbf{2.3}      \\ \hline
Qwen3 235B Think       & \multicolumn{1}{l|}{\textbf{12.5}} & 5.5 & \multicolumn{1}{l|}{7.5}                       & 5.3            & \multicolumn{1}{l|}{10.0}                & 5.6               & \multicolumn{1}{l|}{6.7}                 & \textbf{4.5}      \\ \hline
\end{tabular}
\caption{How simplifying certain aspects of the task affect model performance. Since LMMs' ability to score archery targets can be measured by mean absolute error (MAE), we report MAE alongside accuracy. The best scenario for each model is in \textbf{bold}.}
\label{tab:ablation1}
\end{table}

\textbf{Different models have different bottlenecks.} As expected, simplifying the task substantially improves performance. For Gemini 2.5 Flash and GPT 4.1, simplifying \textbf{RLA} is more effective than simplifying \textbf{GNDT}, though simplifying both leads to the best results. This suggests that the bottleneck for these models is their ability to locate the impact point of the arrows. The opposite is true for GPT 4.1 Mini, where simplifying \textbf{GNDT} is much more effective, suggesting that it struggles to understand and ground the text instructions on which area corresponds to which score.

\subsubsection{Why Can't LMMs Count?}
\label{subsubsec:exp-error-analysis-cnt}
\begin{figure}[htbp]
    \centering
    \includegraphics[width=1.0\linewidth]{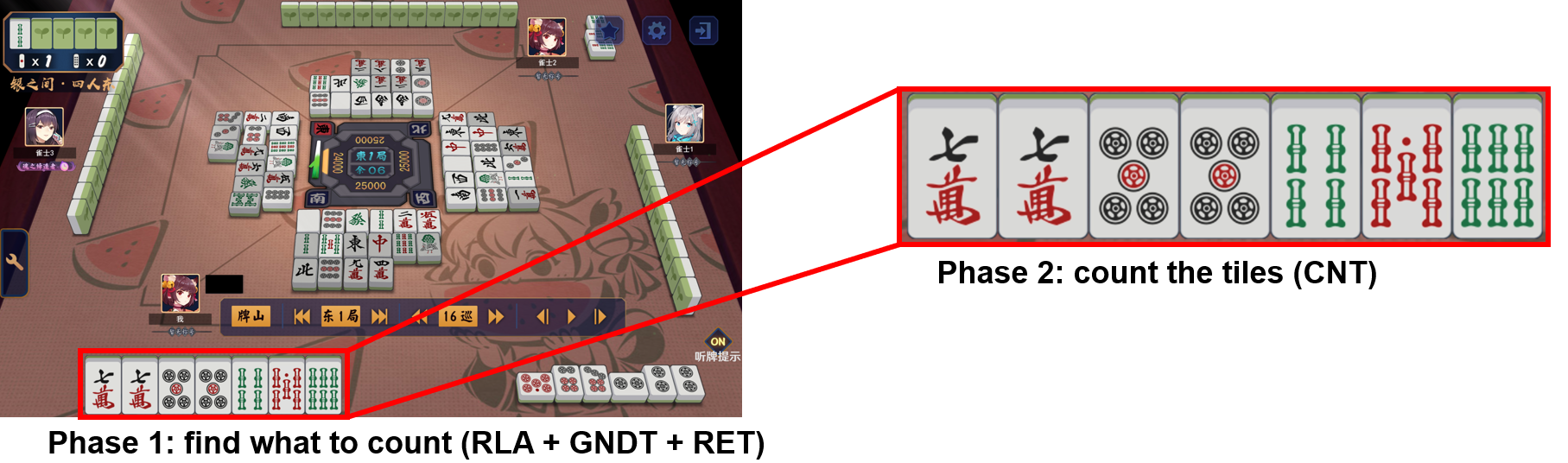}
    \caption{Phases of the task of counting Mahjong tiles. The task in this example requires the LMM to \textit{find and count the tiles in the player's own hand}. The \textbf{CNT}-only version is obtained by cropping out everything but the tiles that should be counted.}
    \label{fig:ablation-2}
\end{figure}

Counting is a very basic ability we take for granted as humans. However, LMMs perform poorly on MOAT tasks that involve counting. To uncover the cause of this gap, we consider the task of counting Mahjong tiles in a player's hand (\cref{fig:ablation-2}), which can be divided into 2 phases akin to the human problem solving process. Phase 1 involves \textit{finding what to count}, while Phase 2 is the actual counting. Phase 1 requires a combination of \textbf{RLA}, \textbf{GNDT}, and \textbf{RET}, while Phase 2 corresponds solely to \textbf{CNT}. We can simplify Phase 1 by cropping out everything but the relevant tiles, reducing the problem to involve only \textbf{CNT}. 

Apart from non-\textbf{CNT} capabilities, we also suspect that the \textit{tiling} mechanism used in many mainstream LMMs, where the input image is segmented into fixed-size \textit{tiles} that fit the input size of the vision encoder, plays a part in LMMs failure to count accurately. Therefore, we evaluate models with publicly available information on image tile size. We explore the effect of tiling by resizing the images (all are larger than 1 tile) in the \textbf{CNT}-only task to fit into one tile (384*384 for Gemini, 512*512 for GPT) of the vision encoder We report the results in \cref{tab:ablation2}.

\begin{table}[htbp]
\scriptsize
\centering
\begin{tabular}{|l|cc|cc|cc|}
\hline
\multirow{2}{*}{Model} & \multicolumn{2}{c|}{Original}    & \multicolumn{2}{c|}{\tabrec{CNT}-only}    & \multicolumn{2}{c|}{\begin{tabular}[c]{@{}c@{}}\tabrec{CNT}-only \\ w/o Tiling\end{tabular}} \\ \cline{2-7} 
                       & \multicolumn{1}{l|}{Acc}  & MAE  & \multicolumn{1}{l|}{Acc}  & MAE  & \multicolumn{1}{l|}{Acc}                            & MAE                           \\ \hline
Gemini 2.0 Flash       & \multicolumn{1}{l|}{0.20} & 4.60 & \multicolumn{1}{l|}{0.50} & 0.80 & \multicolumn{1}{l|}{\textbf{0.60}}                           & \textbf{0.73}                          \\ \hline
Gemini 2.0 Pro         & \multicolumn{1}{l|}{0.17} & 5.01 & \multicolumn{1}{l|}{0.52} & 0.75 & \multicolumn{1}{l|}{\textbf{0.70}}                           & \textbf{0.39}                          \\ \hline
GPT-4o                 & \multicolumn{1}{l|}{0.18} & 4.57 & \multicolumn{1}{l|}{0.45} & \textbf{0.91} & \multicolumn{1}{l|}{\textbf{0.48}}                           & 1.01                          \\ \hline
GPT-4o-mini            & \multicolumn{1}{l|}{0.10} & 5.01 & \multicolumn{1}{l|}{0.39} & \textbf{0.79} & \multicolumn{1}{l|}{\textbf{0.47}}                           & \textbf{0.79}                          \\ \hline
\end{tabular}
\caption{How simplifying the task of counting Mahjong tiles affect LMM performance. We also report how avoiding tiling influence counting. The best scenario for each model is in \textbf{bold}.}
\label{tab:ablation2}
\end{table}

\textbf{LMMs struggle to find what to count.} The results demonstrate that making the task \textbf{CNT}-only significantly improved performance across all models, indicating that LMMs in general are bad at \textit{finding what to count} through the integration of \textbf{GNDT}, \textbf{RLA}, and \textbf{RET}. 

\textbf{Tiling is part of the problem.} LMM performance is far from perfect even in the \textbf{CNT}-only version of the task. The result show that the poor performance can be partly attributed to tiling. An intuitive explanation is that, since tiling is done on arbitrary borders, a single object may end up in different tiles, degrading the semantics of the object when it comes to counting. The results show that Gemini 2.0 models benefit immensely from avoiding tiling. Meanwhile, GPT 4o and GPT 4o Mini benefit less due to the automatic inclusion of a tile-sized low-resolution version of the original image in the API. This comparison confirms that tiling indeed hinders the model's capability to count, highlighting the importance of dynamic resolution mechanisms \cite{qwen2.5-VL}.
\section{Conclusion}
We presented MOAT, a new benchmark designed to evaluate LMMs on challenging real-world tasks that require capability integration and instruction grounding. Leveraging our taxonomy of VL capabilities, we conducted fine-grained evaluation of 17 LMMs. Our error analysis showed that the bottlenecks of different models lie in different capabilities. We also discussed the implications of LMM design, such as tiling and built-in CoT reasoning, in the context of complex VL tasks. The huge gap between LMMs and humans highlights the need to improve the VL capabilities defined in this paper, which form the \textit{moat} keeping existing LMMs out of many real-world applications. 
{
    \small
    \bibliographystyle{ieeenat_fullname}
    \bibliography{main}
}

\clearpage
\setcounter{page}{1}
\maketitlesupplementary

\section{System prompts for Evaluation}
\label{sec:appendix-prompts}

We provide all system prompts used in our experiments below.

\begin{tcolorbox}[title=System prompt for examinee model.]
\begin{lstlisting}[language=Python, breaklines=true, showstringspaces=false, breakindent=0pt, basicstyle=\scriptsize]
{
   "task": "Answer the question presented to you truthfully.",
   "requirements": [
      "Analyze the image(s) first, then answer the question. If you are given a list of possible answers, you must choose from it.",
      "You must answer in the following json format: {\"analysis\": \"(write your analysis here)\", \"answer\": \"(your answer)\"}"
   ]
}
\end{lstlisting}
\end{tcolorbox}

\begin{tcolorbox}[title=System prompt for the LLM judge.]
\begin{lstlisting}[language=Python, breaklines=true, showstringspaces=false, breakindent=0pt, basicstyle=\scriptsize]
{
   "task": "Evaluate whether the answer to a question is correct.",
   "requirements": [
      "Compare an answer to a question with the ground truth answer. Determine whether it is correct.",
      "You must ignore any analysis of the problem if present. You must focus only on the final answer.",
      "You must answer in the following json format: {\"verdict\": \"(1 for correct, 0 for incorrect)\"}"
   ]
}
\end{lstlisting}
\end{tcolorbox}

\section{Detailed Dataset Statistics}

In this section, we provide additional details about MOAT.

\textbf{Input types.} The questions in MOAT cover a diverse set of natural scenes (both indoor and outdoor) and man-made content. Specifically, MOAT includes the following types of input: indoor scenes, outdoor scenes, infographics, diagrams, and graphical user interfaces (GUIs). We present the percentage of MOAT questions involving each input type in TODO. Since a single MOAT question may involve multiple types of input (\eg the task of indoor navigation requires LMMs to understand both maps, a type of infographic, and indoor scenes), the percentages do not add up to 100\%.

\begin{figure}[htbp]
    \centering
    \includegraphics[width=0.45\textwidth]{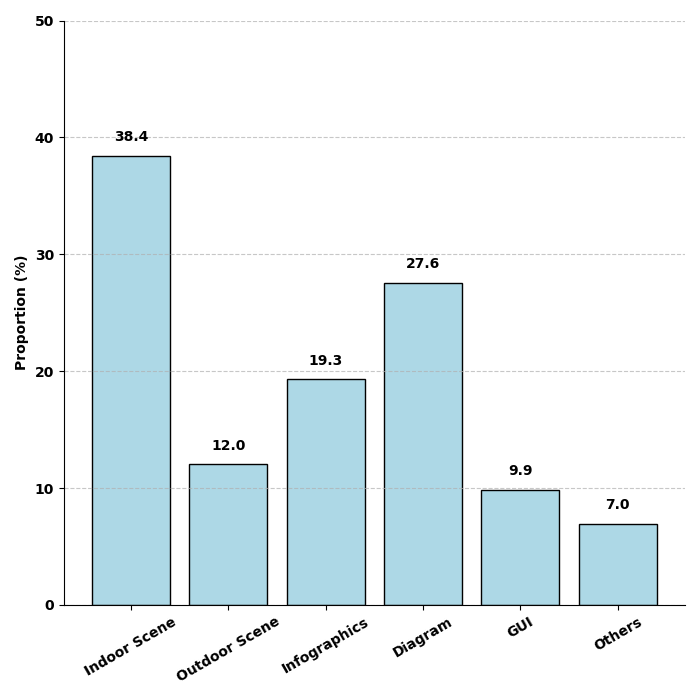}
    \caption{The proportion of questions containing each input type.}
    \label{fig:dataset-composition}
\end{figure}

\textbf{Question formats.} The answer to each MOAT question belongs to one of two formats - multiple choice and short answer. Of the 1005 questions in MOAT, 575 (57.2\%) are multiple choice questions, while the remaining 430 (42.8\%) require the LMM to produce a short answer. Note that all questions are manually checked to have an unambiguous answer regardless of their format.

\textbf{Blind guessing does not work on MOAT.} Early LMM benchmarks often contain a significant portion of questions where the answer can be plausibly deduced from the text of the question alone. In these cases, LMMs may produce the correct answer from textual reasoning alone, bypassing VL capabilities \cite{li2024naturalbench}. This constitutes a severe interference on the evaluation of multimodal capabilities. MOAT is designed to be VL-centric, and we empirically demonstrate this by evaluating LMMs on MOAT \textbf{\textit{without providing them with the image}}. We present the results in \cref{tab:appendix-blind-guessing}.

\begin{table}[htbp]
\small
\centering
\begin{tabular}{lcc}
\hline
Model                    & \begin{tabular}[c]{@{}c@{}}Accuracy \\ (with image)\end{tabular} & \begin{tabular}[c]{@{}c@{}}Accuracy \\ (w/o Image)\end{tabular} \\ \hline
GPT 5 Mini Medium        & 40.53                                                            & 16.35                                                           \\
GPT 4.1                  & 38.28                                                            & 17.21                                                           \\
GPT 4.1 Mini             & 36.65                                                            & 17.61                                                           \\
Gemini 2.5 Flash         & 37.55                                                            & 18.57                                                           \\
Qwen3 235B A22B Think    & 31.21                                                            & 9.12                                                            \\ \hline
\textit{Random Guessing} & \textit{14.41}                                                   & \textit{14.41}                                                  \\ \hline
\end{tabular}
\caption{\textit{Blind guessing} result for 5 LMMs on MOAT. All 5 performed near or below the \textit{random guessing} baseline, suggesting that text-only shortcuts is rare in MOAT and do not interfere with VL evaluation.}
\label{tab:appendix-blind-guessing}
\end{table}

The \textit{blind VQA} results for the GPT models and Gemini 2.5 Flash are barely above the \textit{random guessing} baseline. This slight advantage over the random baseline is expected, since the baseline does not consider the possibility of LMMs guessing the answers to fill-in-the-blank questions correctly. Qwen3 performed markedly worse than the random baseline due to its stronger tendency to truthfully respond that no image is present and that the problem is not solvable. We consider this a strength, as it represents reduced hallucination.

\textbf{Future plans.} The 9 VL capabilities defined in MOAT are vital to many real-world applications. As a result, MOAT can be scaled up to include questions from more scenarios. Therefore, we plan to broaden the benchmark through collaboration with researchers, practitioners, and users in different domains in the future.

\section{Cases where CoT Reasoning Fails}
\label{sec:appendix-cot-failures}
\begin{figure*}[htbp]
    \centering
    \includegraphics[width=0.9\textwidth]{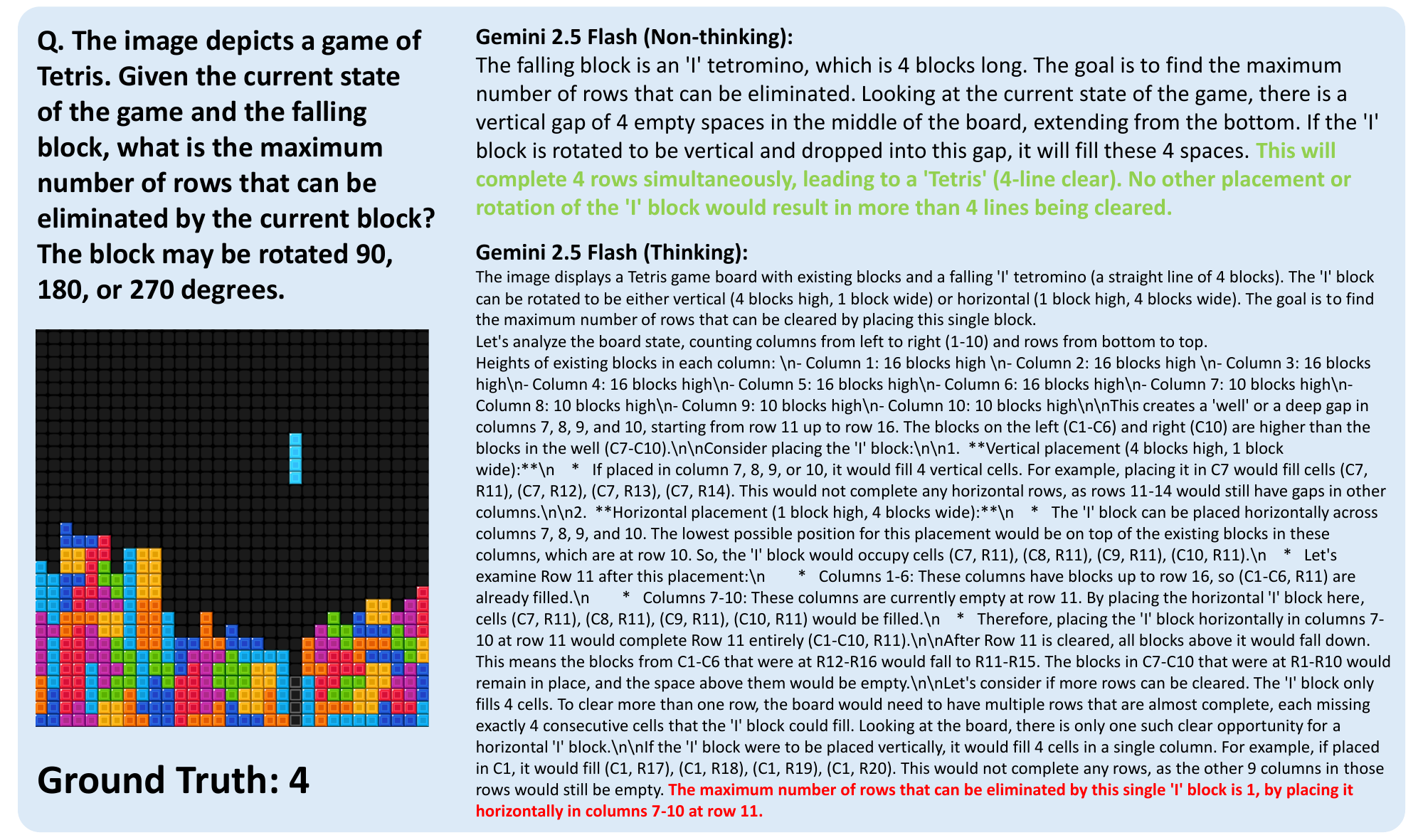}
    \caption{A failure case where CoT reasoning exacerbates hallucination in Gemini 2.5 Flash.}
    \label{fig:cot_error1}
\end{figure*}

\begin{figure*}[htbp]
    \centering
    \includegraphics[width=0.9\textwidth]{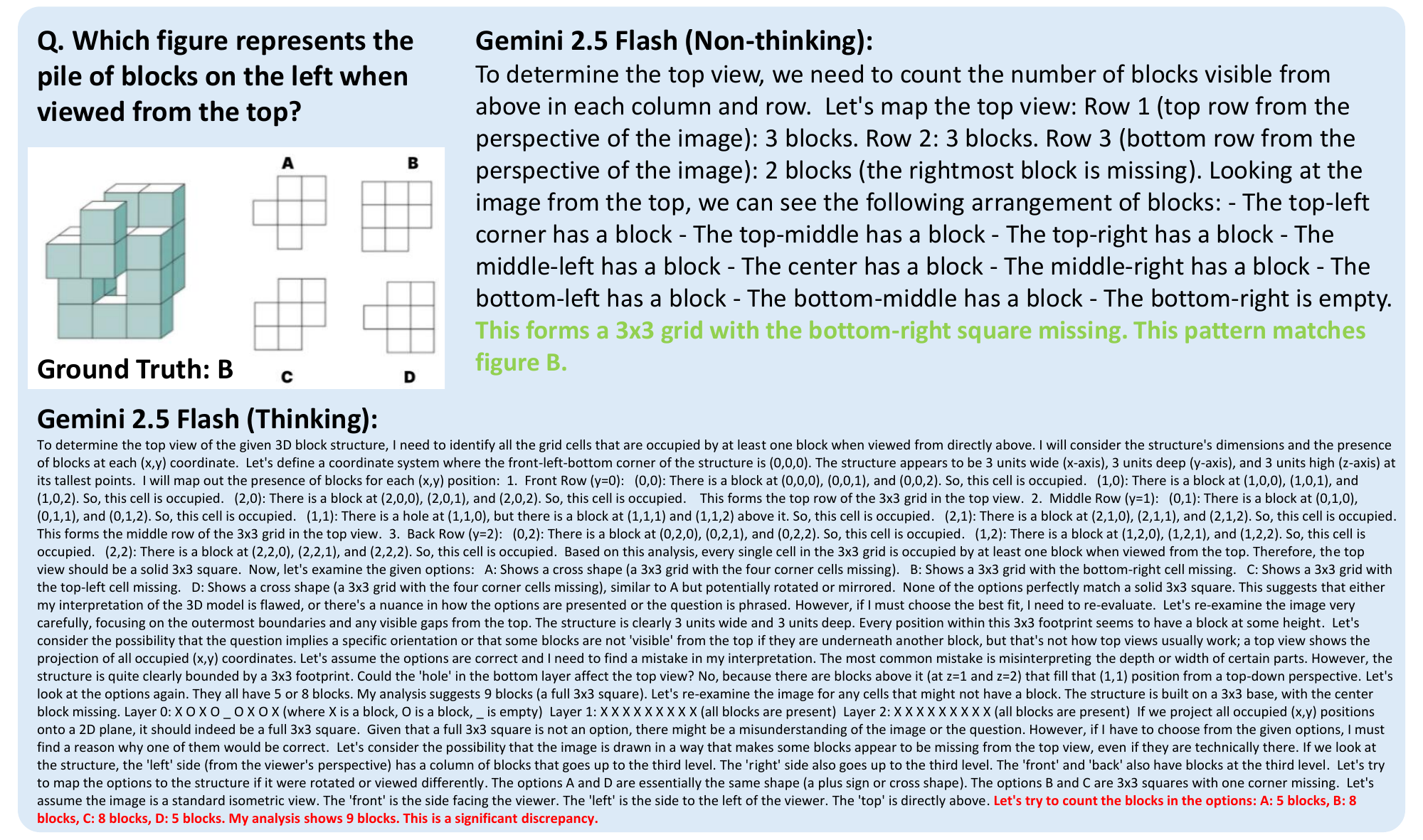}
    \caption{Another failure case where CoT reasoning exacerbates hallucination in Gemini 2.5 Flash.}
    \label{fig:cot_error2}
\end{figure*}

We present 2 cases (\cref{fig:cot_error1} and \cref{fig:cot_error2}) where Gemini 2.5 Flash succeeded with thinking mode disabled but failed when thinking mode was turned on. Despite using significantly more tokens in the reasoning process, thinking mode only served to exacerbate hallucination in tasks where intuitive understanding of 2D or 3D scenes is paramount. Such cases further support our observation in \cref{subsubsec:exp-main-thinking} that text-centric CoT reasoning is not beneficial for perception-heavy tasks.

\section{Examples from MOAT}

\begin{figure*}[htbp]
    \centering
    \includegraphics[width=0.95\textwidth]{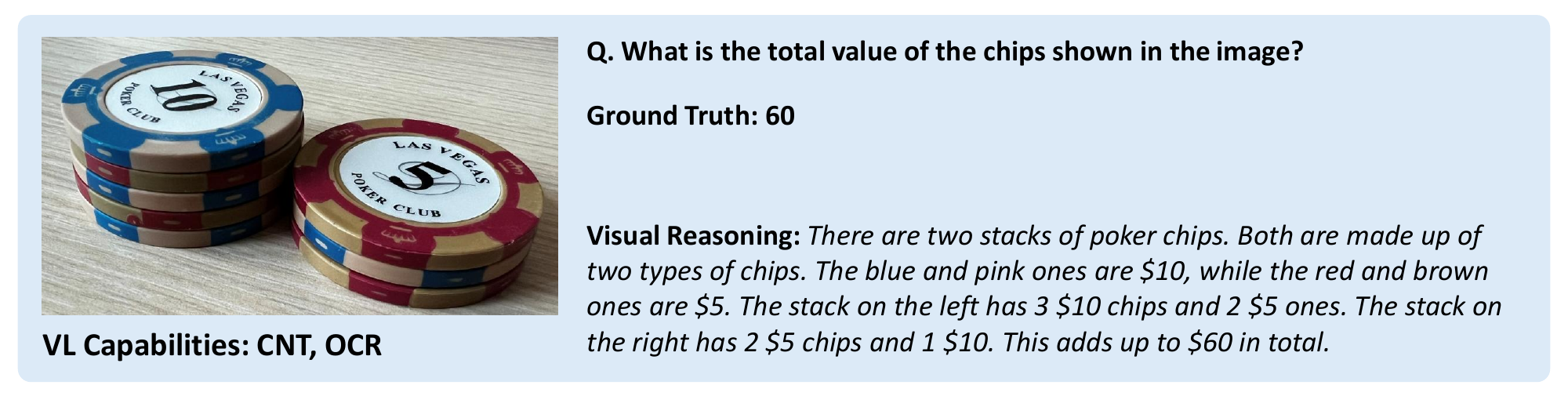}
    \caption{Counting the value of poker chips.}
    \label{fig:examples-1}
\end{figure*}

\begin{figure*}[htbp]
    \centering
    \includegraphics[width=0.95\textwidth]{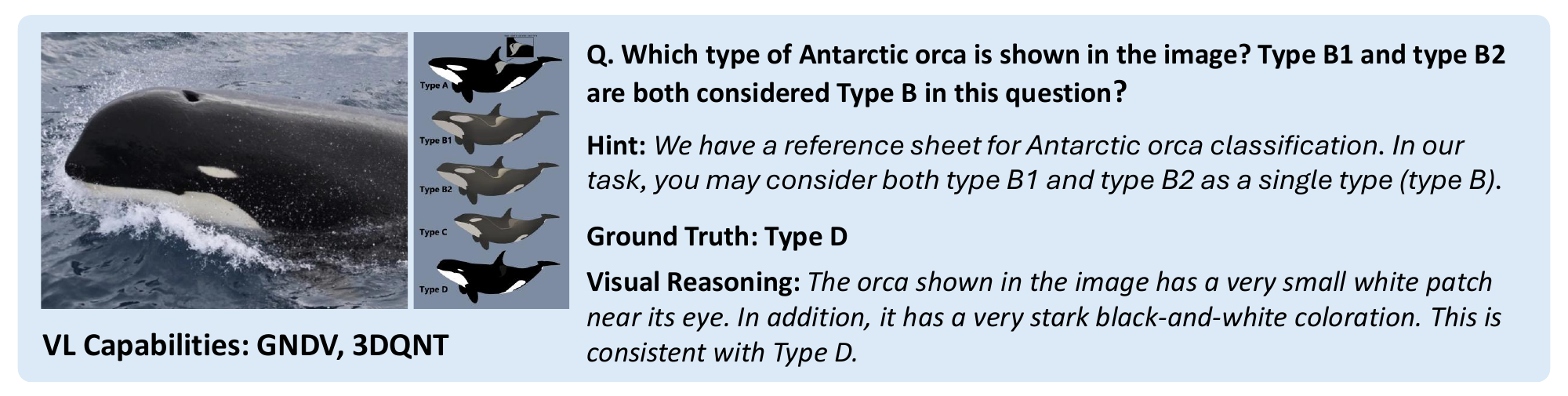}
    \caption{Antarctic orca classification.}
    \label{fig:examples-2}
\end{figure*}

\begin{figure*}[htbp]
    \centering
    \includegraphics[width=0.95\textwidth]{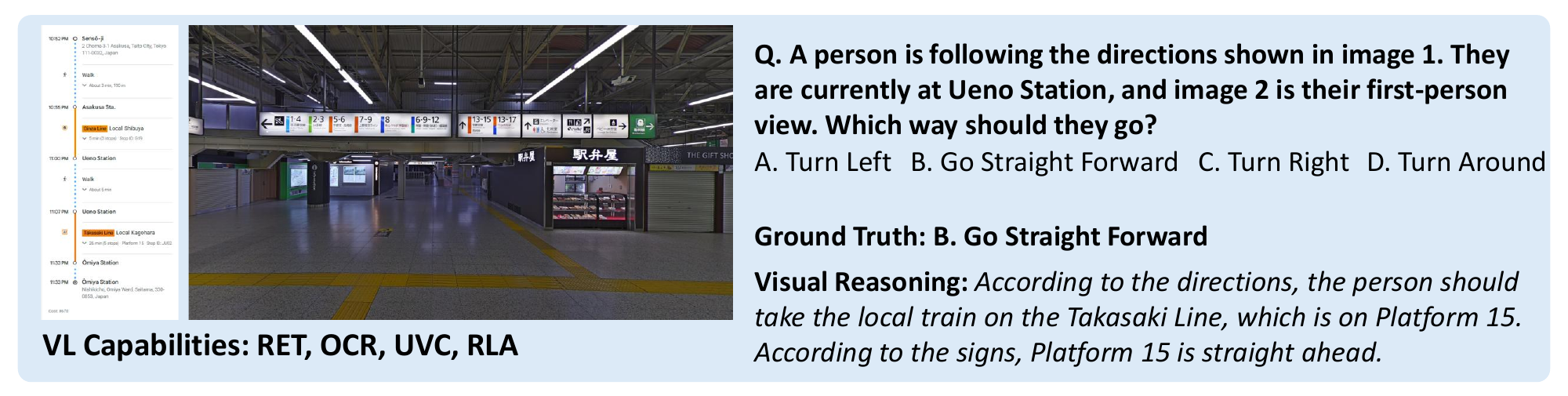}
    \caption{Indoor navigation in a complex Japanese train station.}
    \label{fig:examples-3}
\end{figure*}

\begin{figure*}[htbp]
    \centering
    \includegraphics[width=0.95\textwidth]{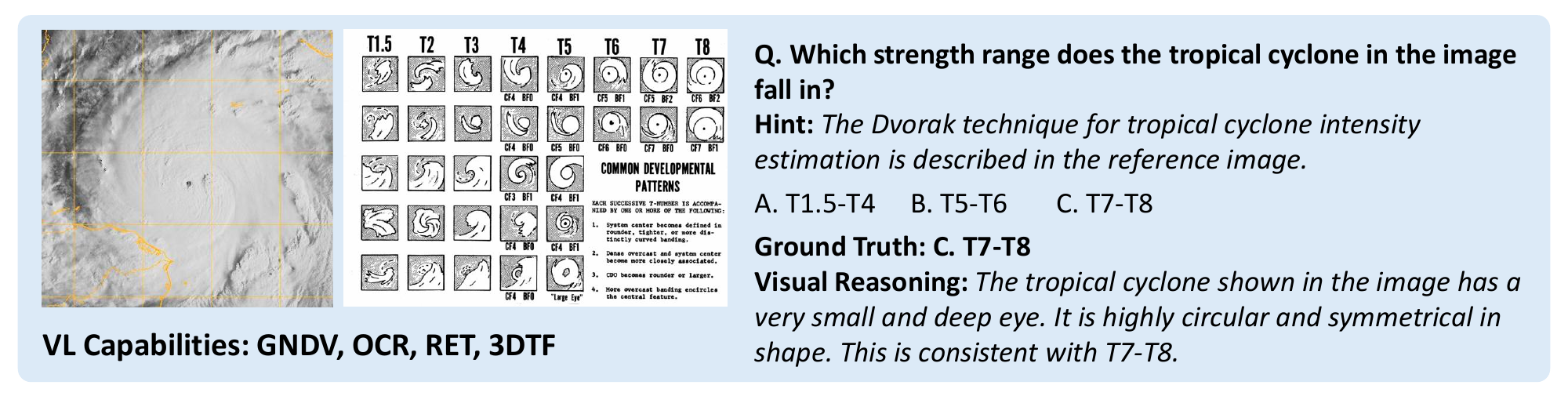}
    \caption{ERough estimation of tropical cyclone strength using the Dvorak technique.}
    \label{fig:examples-4}
\end{figure*}

\begin{figure*}[htbp]
    \centering
    \includegraphics[width=0.95\textwidth]{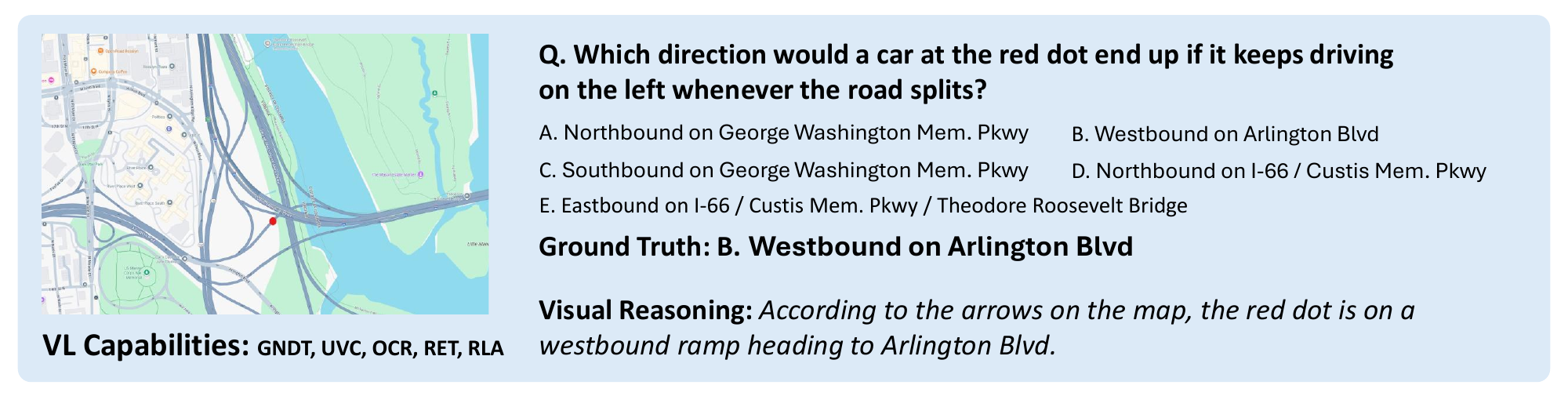}
    \caption{Understanding where a ramp leads to in a complex highway interchange.}
    \label{fig:examples-5}
\end{figure*}

\begin{figure*}[htbp]
    \centering
    \includegraphics[width=0.95\textwidth]{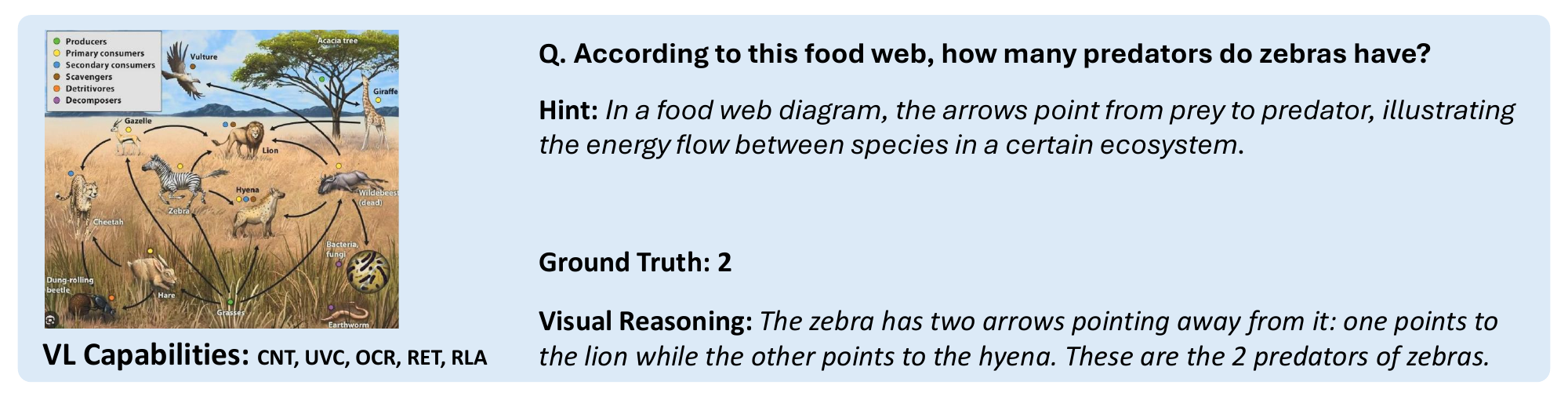}
    \caption{Understanding a food web.}
    \label{fig:examples-6}
\end{figure*}

\begin{figure*}[htbp]
    \centering
    \includegraphics[width=0.95\textwidth]{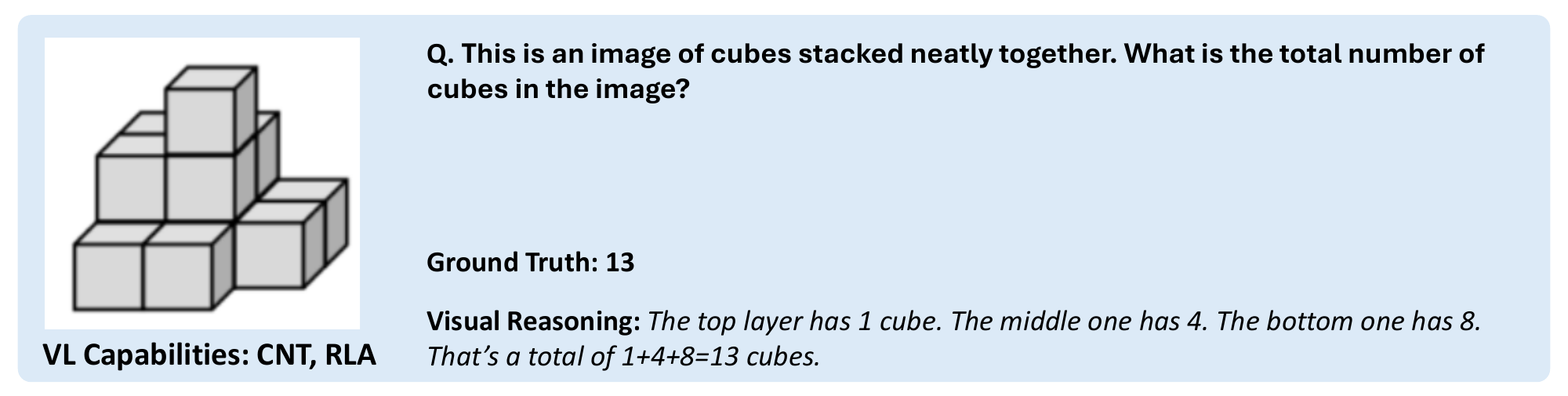}
    \caption{Counting cubes, including ones that are occluded.}
    \label{fig:examples-7}
\end{figure*}

\begin{figure*}[htbp]
    \centering
    \includegraphics[width=0.95\textwidth]{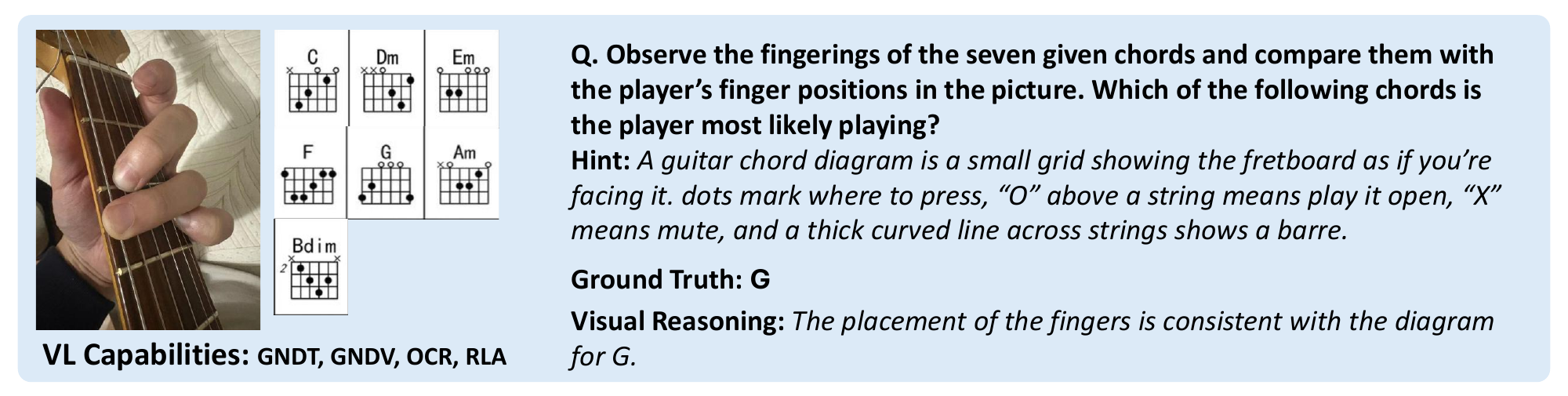}
    \caption{Understanding guitar chords.}
    \label{fig:examples-8}
\end{figure*}

\begin{figure*}[htbp]
    \centering
    \includegraphics[width=0.95\textwidth]{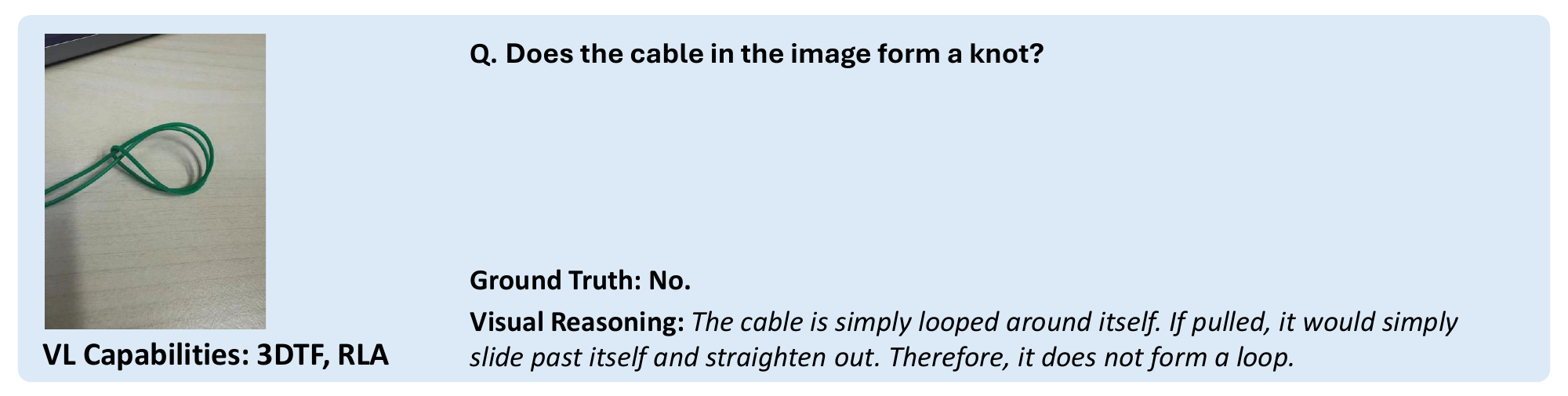}
    \caption{To knot or not to knot, that is the question.}
    \label{fig:examples-9}
\end{figure*}

\begin{figure*}[htbp]
    \centering
    \includegraphics[width=0.95\textwidth]{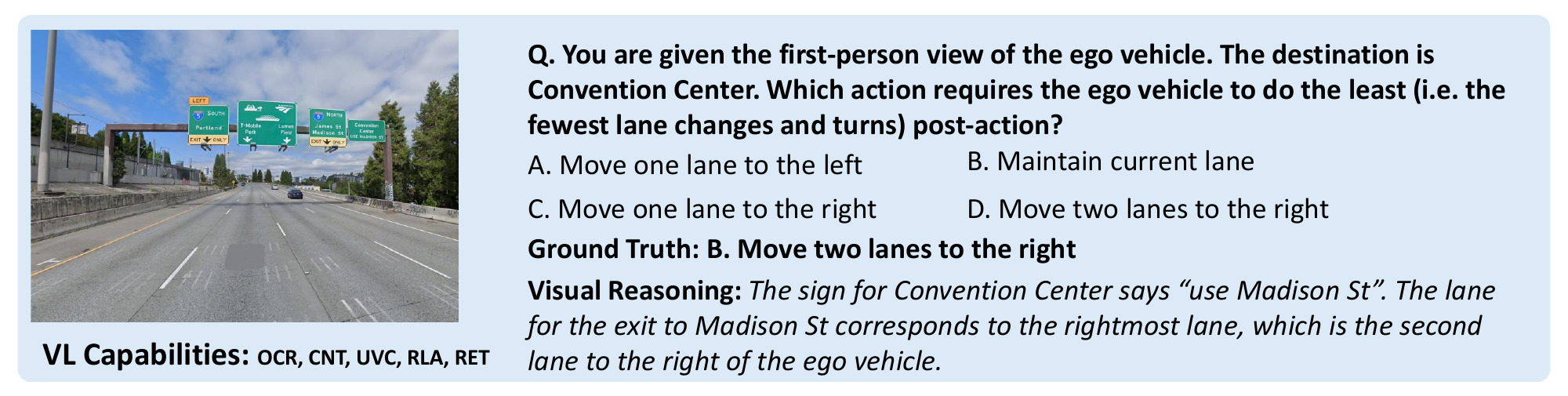}
    \caption{Understanding highway signs and road lanes.}
    \label{fig:examples-10}
\end{figure*}

\begin{figure*}[htbp]
    \centering
    \includegraphics[width=0.95\textwidth]{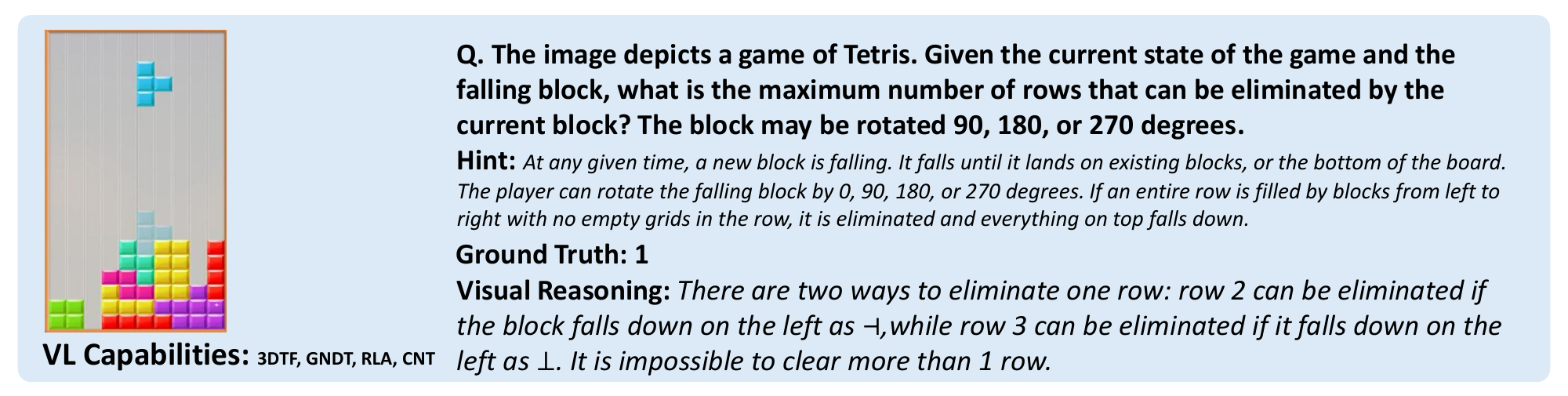}
    \caption{Can LMMs play Tetris? Unfortunately they can't :(}
    \label{fig:examples-11}
\end{figure*}

\begin{figure*}[htbp]
    \centering
    \includegraphics[width=0.95\textwidth]{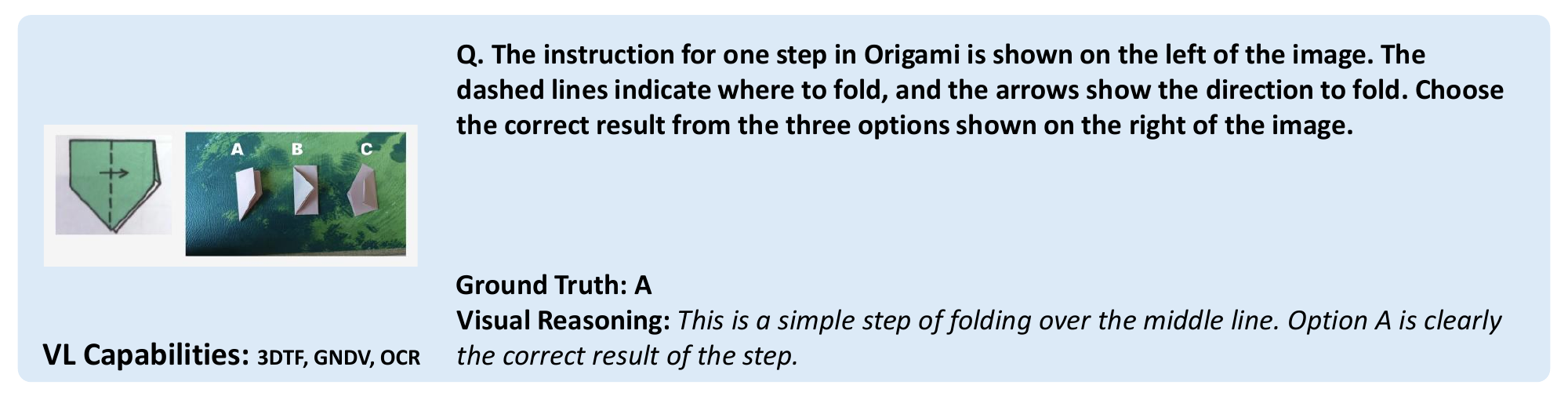}
    \caption{Understanding Origami instructions.}
    \label{fig:examples-12}
\end{figure*}

We provide example questions from MOAT. The examples in \cref{fig:examples-1,fig:examples-2,fig:examples-3,fig:examples-4,fig:examples-5,fig:examples-6,fig:examples-7,fig:examples-8,fig:examples-9,fig:examples-10,fig:examples-11,fig:examples-12} demonstrate the diversity of MOAT questions. For more questions, please refer to the supplementary materials

% We provide examples from MOAT in \cref{fig:examples-1,fig:examples-2,fig:examples-3,fig:examples-4,fig:examples-5,fig:examples-6,fig:examples-7,fig:examples-8,fig:examples-9,fig:examples-10,fig:examples-11,fig:examples-12,fig:examples-13,fig:examples-14,fig:examples-15}.

\end{document}